\documentclass[journal]{IEEEtran}

\ifCLASSINFOpdf
\else
\fi

\usepackage{times}
\usepackage{epsfig}
\usepackage{graphicx}
\usepackage{amsmath}
\usepackage{amssymb}
\usepackage{graphicx}
\usepackage{cite}
\usepackage{enumerate}
\usepackage{enumitem}
\usepackage{cases}
\usepackage{multirow}
\usepackage{verbatim}
\usepackage{amssymb}
\usepackage{CJK}
\usepackage{algorithm}
\usepackage{algorithmicx}
\usepackage{algpseudocode}

\usepackage{color}
\usepackage{bm}
\usepackage{booktabs}
\usepackage{makecell}
\usepackage{url}
\usepackage{hyperref}

\begin{document}

\title{Review of Visual Saliency Detection with\\ Comprehensive Information}


\author{Runmin Cong, Jianjun Lei,~\IEEEmembership{Senior Member,~IEEE,} Huazhu Fu,~\IEEEmembership{Senior Member,~IEEE,}\\ Ming-Ming Cheng, Weisi Lin,~\IEEEmembership{Fellow,~IEEE,} and Qingming Huang,~\IEEEmembership{Fellow,~IEEE}
\thanks{Manuscript received March 08. 2018; revised July 25, 2018 and August 28, 2018; accepted September 03, 2018. This work was supported in part by the National Natural Science Foundation of China under Grant 61722112, Grant 61520106002, Grant 61731003, Grant 61332016, Grant 61620106009, Grant U1636214, Grant 61602344, and in part by the National Key Research and Development Program of China under Grant 2017YFB1002900. (\emph{Corresponding author: Jianjun Lei})}
\thanks{R. Cong and J. Lei are with the School of Electrical and Information Engineering, Tianjin University, Tianjin 300072, China (e-mail: rmcong@tju.edu.cn; jjlei@tju.edu.cn).}
\thanks{H. Fu is with the Inception Institute of Artificial Intelligence, Abu Dhabi, United Arab Emirates (e-mail: huazhufu@gmail.com).}
\thanks{M.-M. Cheng is with the School of Computer and Control Engineering, Nankai University, Tianjin 300071, China (e-mail: cmm@nankai.edu.cn)}
\thanks{W. Lin is with the School of Computer Science Engineering, Nanyang Technological University, Singapore 639798. (e-mail: wslin@ntu.edu.sg).}
\thanks{Q. Huang is with the School of Computer and Control Engineering, University of Chinese Academy of Sciences, Beijing 100190, China (e-mail: qmhuang@ucas.ac.cn).}
\thanks{Copyright 20xx IEEE. Personal use of this material is permitted. However, permission to use this material for any other purposes must be obtained from the IEEE by sending an email to pubs-permissions@ieee.org.}}

\markboth{IEEE Transactions on Circuits and Systems for Video Technology, ~Vol.~xx, No.~xx, xxxx~2018}%
{Shell \MakeLowercase{\textit{et al.}}: Bare Demo of IEEEtran.cls for IEEE Journals}

\maketitle

\begin{abstract}
Visual saliency detection model simulates the human visual system to perceive the scene, and has been widely used in many vision tasks. With the development of acquisition technology, more comprehensive information, such as depth cue, inter-image correspondence, or temporal relationship, is available to extend image saliency detection to RGBD saliency detection, co-saliency detection, or video saliency detection. RGBD saliency detection model focuses on extracting the salient regions from RGBD images by combining the depth information. Co-saliency detection model introduces the inter-image correspondence constraint to discover the common salient object in an image group. The goal of video saliency detection model is to locate the motion-related salient object in video sequences, which considers the motion cue and spatiotemporal constraint jointly. In this paper, we review different types of saliency detection algorithms, summarize the important issues of the existing methods, and discuss the existent problems and future works. Moreover, the evaluation datasets and quantitative measurements are briefly introduced, and the experimental analysis and discission are conducted to provide a holistic overview of different saliency detection methods.
\end{abstract}


\IEEEpeerreviewmaketitle

\section{Introduction}

\IEEEPARstart{H}{UMAN} visual system works as a filter to allocate more attention to the attractive and interesting regions or objects for further processing. Humans can exhibit visual fixation, which is maintaining of the visual gaze on a single location. Inspired by this visual perception phenomena, some visual saliency models focus on predicting human fixations \cite{R0}. In addition, driven by computer vision applications, some visual saliency models aim at identifying the salient regions from the image or video \cite{R92}. In this survey, we mainly review the latest progress of salient object detection, which has been applied in image/video segmentation \cite{R1,R2}, image/video retrieval \cite{R3,R4}, image retargeting \cite{R5,R6}, image compression \cite{R7}, image enhancement \cite{R8,R9-2,R9}, video coding \cite{R10}, foreground annotation \cite{R11}, quality assessment \cite{R12,R13}, thumbnail creation \cite{R14}, action recognition \cite{R15}, and video summarization \cite{R16}.\par

The last decade has witnessed the remarkable progress of image saliency detection, and a plenty of methods have been proposed and achieved the superior performances, especially the deep learning based methods have yielded a qualitative leap in performances. Following \cite{R92}, image saliency detection methods can be classified into  bottom-up model \cite{R18,R19,R20,R21,R22,R23,R24,R25,R26,R27,R28,R29,R93,R95} and top-down model \cite{R21-2,R21-3,R21-4,R30,R31,R32,R33,R34,R35,R36,R37,R38,R38-2,R96,R130,R131}. Bottom-up model is stimulus-driven, which focuses on exploring low-level vision features. Some visual priors are utilized to describe the properties of salient object based on the visual inspirations from the human visual system, such as contrast prior \cite{R18}, background prior \cite{R22,R24,R29}, and compactness prior \cite{R26}. In addition, some traditional techniques are also introduced to achieve image saliency detection, such as frequency domain analysis \cite{R17}, sparse representation \cite{R20}, cellular automata \cite{R23}, random walks \cite{R24}, low-rank recovery \cite{R27}, and Bayesian theory \cite{R28}. Top-down model is task-driven, which utilizes supervised learning with labels and achieves high performance. Especially, deep learning technique has been demonstrated the powerful ability in saliency detection. Some hierarchical deep networks for saliency detection are proposed, such as SuperCNN \cite{R31}, and DHSNet \cite{R34}. In addition, the multi-scale or multi-context deep saliency network is proposed to learn more comprehensive features, such as deep contrast network \cite{R33}, multi-context deep learning framework \cite{R38-2}, multi-scale deep network \cite{R96}, and network with short connections \cite{R36}. The symmetrical network is also introduced in saliency detection, such as the encoder-decoder fully cnvolutional networks \cite{R37}. Moreover, some deep weakly supervised methods for salient object detection are proposed by using the image-level supervision \cite{R130} or noisy annotation \cite{R131}. \par

\begin{figure}[!t]
\centering
\includegraphics[width=1\linewidth]{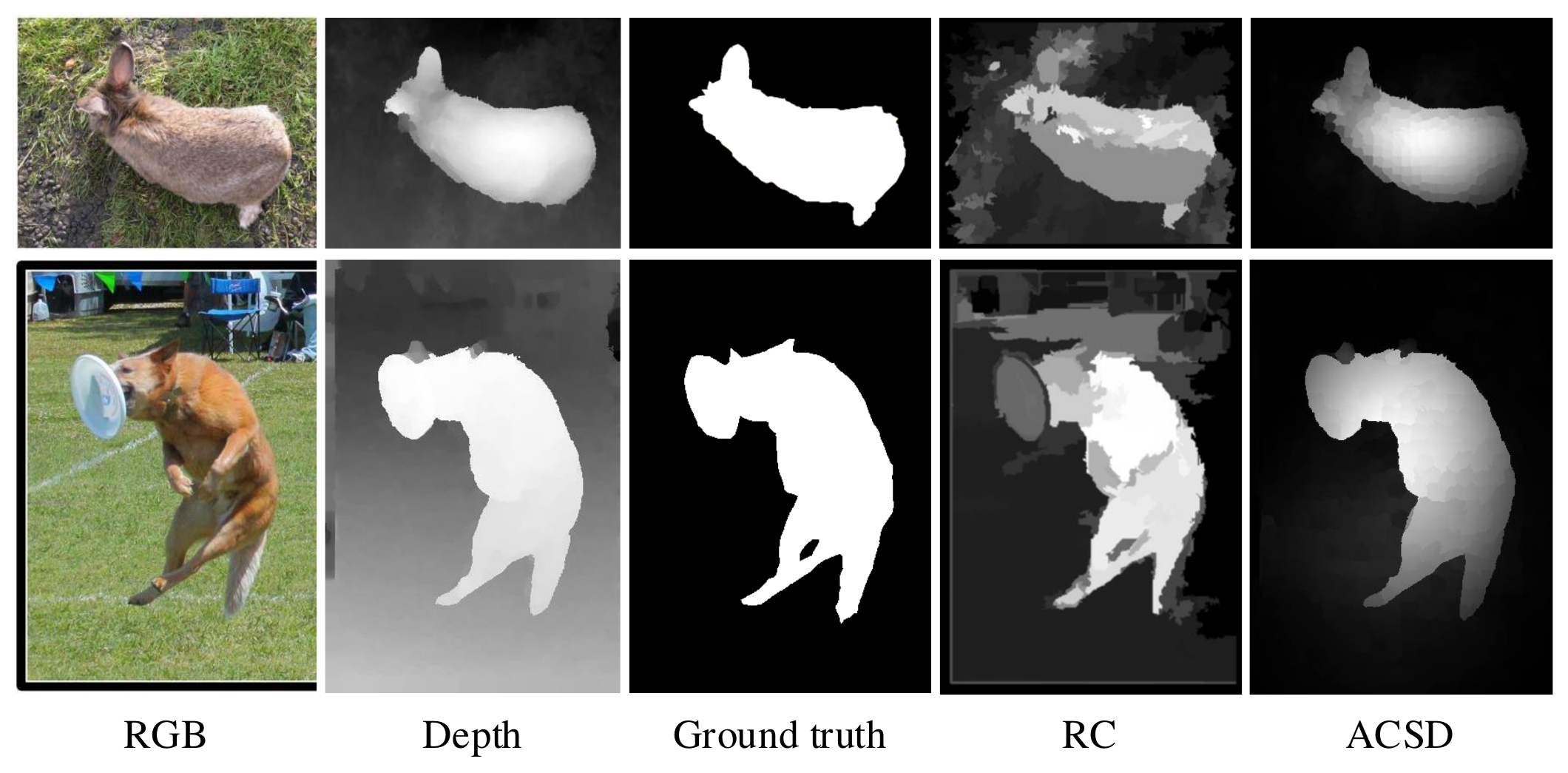}
\caption{Some illustrations of saliency detection with and without depth cue. The first three columns shows the RGB image, depth map, and ground truth. The fourth column shows the image saliency detection result using the RC method \cite{R18}. The fifth column represents the RGBD saliency detection result using the ACSD method [64].}
\label{fig1}
\end{figure}

In fact, the human visual system can not only perceive the appearance of the object, but also be affected by the depth information from the scene. With the development of imaging devices, the depth map can be acquired conveniently, which lays the data foundation for RGBD saliency detection \cite{R39-2}. Generally, there are three options for 3D depth imaging, i.e., structured light \cite{R119}, TOF (Time-of-Flight) \cite{R120}, and binocular imaging \cite{R121}. The structured light pattern (e.g., Kinect) captures the depth information via the change of light signal projected by the camera, which can obtain high-resolution depth map. The TOF system (e.g., Camcube) estimates the depth through the round-trip time of the light pulses, which has good anti-jamming performance and wider viewing angle. The stereo imaging system takes photo pair via stereo camera and calculates the object's disparity based on two-view geometry. Depth map can provide many useful attributes for foreground extraction from the complex background, such as shape, contour, and surface normal. Some examples of saliency detection with and without depth cue are shown in Fig. \ref{fig1}. As can be seen, utilizing the depth cue, RGBD saliency model achieves superior performance with consistent foreground enhancement. However, how to effectively exploit the depth information to enhance the identification of salient object has not yet reached a consensus, and still needs to be further investigated. Considering the ways of using depth information, we divide the RGBD saliency detection model into depth feature based method \cite{R39,R40,R41,R42,R51,R52,R52-2,R52-3} and depth measure based method \cite{R43,R44,R45,R46,R47,R48,R50,R49}. Depth feature based method focuses on taking the depth information as a supplement to color feature, and depth measure based method aims at capturing comprehensive attributes from the depth map (e.g., shape) through the designed depth measurements.\par

\begin{figure}[!t]
\centering
\includegraphics[width=1\linewidth]{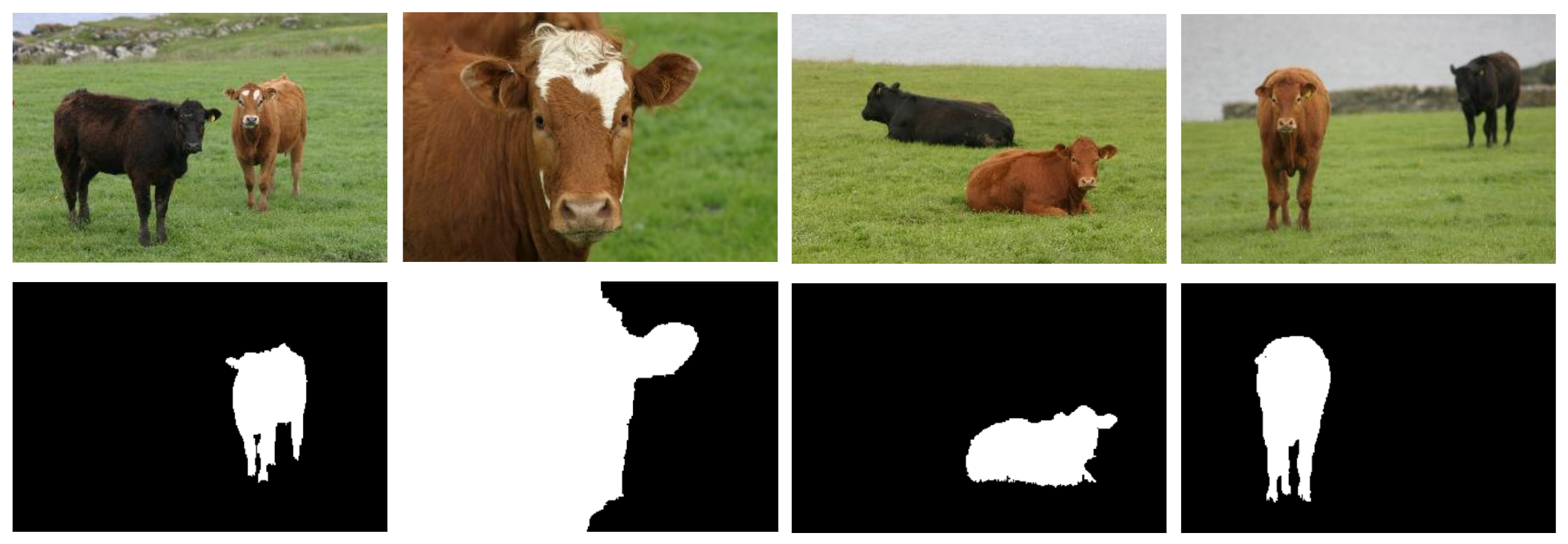}
\caption{Examples of the co-saliency detection model on the iCoseg dataset. The first row presents the input images, and the second row shows the ground truth for co-saliency detection.}
\label{fig2-1}
\end{figure}

In recent years, with the explosive growth of data volume, human need to process multiple relevant images collaboratively. As an emerging and challenging issue, co-saliency detection gains more and more attention from researchers, which aims at detecting the common and salient regions from an image group containing multiple related images, while the categories, intrinsic attributes, and locations are entirely unknown \cite{R118}. In general, three properties should be satisfied by the co-salient object, i.e., (1) the object should be salient in each individual image, (2) the object should be repeated in most of the images, and (3) the object should be similar in appearance among multiple images. Some visual examples of co-saliency detection are provided in Fig. \ref{fig2-1}. In the individual image, all the cows should be detected as the salient objects. However, only the brown cow is the common object from the image group. Therefore, the inter-image correspondence among multiple images plays a useful role in representing the common attribute. On the whole, co-saliency detection methods are roughly grouped into two categories according to whether the depth cue is introduced, i.e., RGB co-saliency detection \cite{R53,R54,R55,R57,R58,R60,R61,R132,R56,R59,R62,R63,R64,R65,R66,R67,R68} and RGBD co-saliency detection \cite{R69,R70,R71,R72}. Then, the RGB co-saliency detection methods are further divided into some sub-classes based on different correspondence capturing strategies, i.e., matching based method \cite{R53,R54,R55,R57,R58,R60,R61,R132}, clustering based method \cite{R56}, rank analysis based method \cite{R59,R62}, propagation based method \cite{R63,R64}, and learning based method \cite{R65,R66,R67,R68}.\par

\begin{figure}[!t]
\centering
\includegraphics[width=1\linewidth]{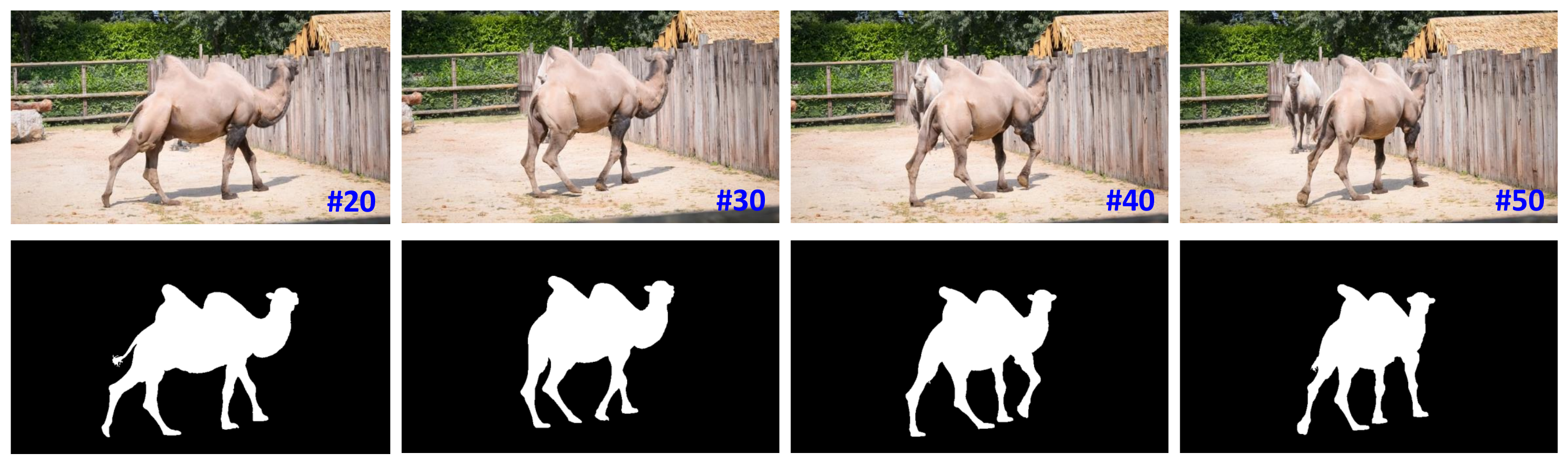}
\caption{Examples of the video saliency detection model on the DAVIS dataset. The first row is the input video frames, and the second row shows the ground truth for video saliency detection.}
\label{fig2-2}
\end{figure}

Different from image data, video sequences contain more abundant appearance information and continuous motion cue, which can better represent the characteristics of the target in a dynamic way. However, the clustered backgrounds, complex motion patterns, and changed views also bring new challenges to interpret video content effectively. Video saliency detection aims at continuously locating the motion-related salient object from the given video sequences by considering the spatial and temporal information jointly. The spatial information represents the intra-frame saliency in the individual frame, while the temporal information provides the inter-frame constraints and motion cues. Fig. \ref{fig2-2} illustrates some examples of video saliency detection. In this camel video, both two camels appeared from 40th frame should be detected as the salient objects through a single image saliency model. However, only the front one is continuously moving and repeating, which is the salient object in this video. The differences between co-saliency detection and video saliency detection lie in two aspects, i.e., (1) The inter-frame correspondence has the temporal property in video saliency detection rather than in co-saliency detection. For co-saliency detection in an image group, the common salient objects have the consistent semantic category, but are not necessarily the same object. By contrast, the salient objects in video are continuous in the time axis and consistent among different frames; (2) In video saliency detection model, motion cue is essential to distinguish the salient object from the complex scene. However, this cue is not included in co-saliency detection model. Similar to the classification strategy of image saliency detection, we divide the video saliency detection methods into two categories, i.e., low-level cue based method \cite{R73,R74,R75,R76,R77,R78,R79,R80,R81,R82,R83,R84,R85} and learning based method \cite{R86,R87,R89,R90,R91}. For clarity, the low-level cue based method is further grouped into fusion model and direct-pipeline model according to feature extraction method, and the learning based method is further divided into supervised method and unsupervised method.\par

\begin{figure}[!t]
\centering
\includegraphics[width=0.9\linewidth]{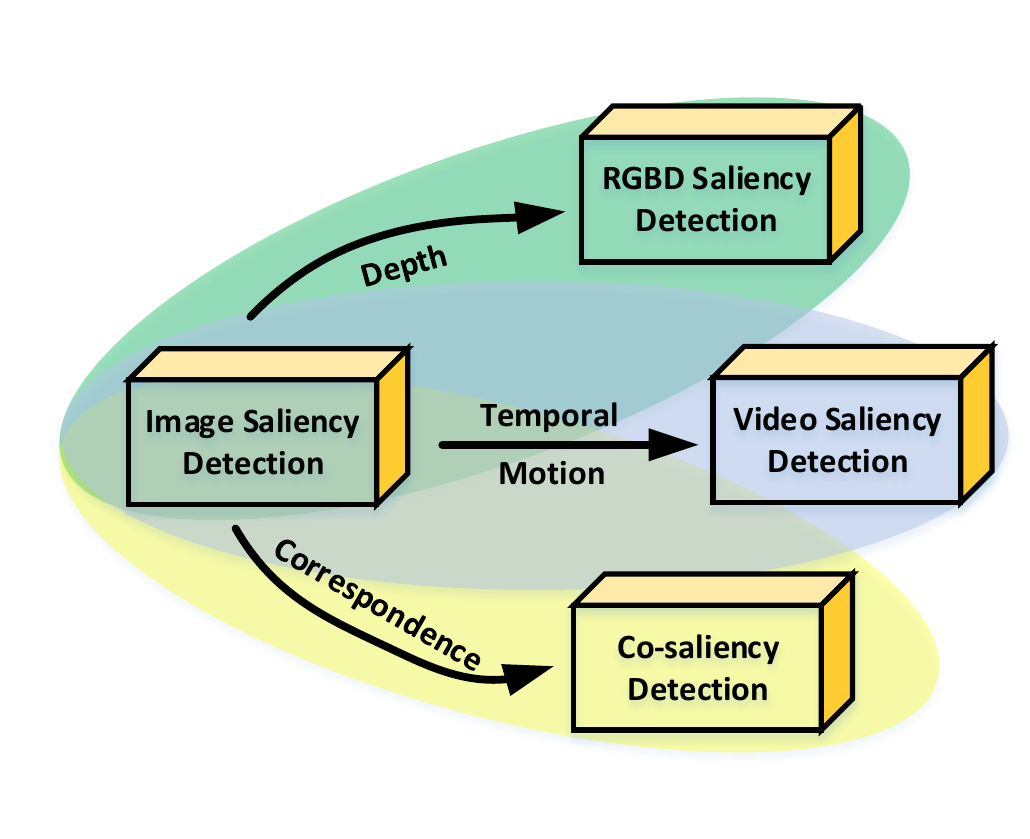}
\caption{Relationships between different visual saliency detection models.
\label{fig3}}
\end{figure}

As stated above, the major relationships among four different visual saliency detection models are summarized in Fig. \ref{fig3}, where the image saliency detection model is the basis for other three models. With the depth cue, RGBD saliency map can be obtained from an image saliency detection model. Introducing the inter-image correspondence, image saliency detection model can be transformed into a co-saliency detection method. Video saliency detection can be derived from an image saliency detection model by combining the temporal correspondence and motion cue, or from a co-saliency detection method by integrating the motion cue. In practice, in order to obtain superior performance, it is necessary to design a specialized algorithm to achieve co-saliency detection or video saliency detection, rather than directly transplanting the image saliency detection algorithms.\par

In this paper, we review different saliency detection models with comprehensive information, and the rest of this paper is organized as follows. Section II surveys the existing RGBD saliency detection models. Section III introduces some co-saliency detection methods. Section IV summarizes the related works of video saliency detection. The experimental comparisons and discussions are presented in Section V. Finally, the conclusion and future work are summarized in Section VI.\par

\section{RGBD Saliency Detection}
Different from image saliency detection, RGBD saliency detection model considers the color information and depth cue together to identify the salient object. As a useful cue for saliency detection, depth information is usually utilized in two ways, i.e., directly incorporating as the feature and designing as the depth measure. Depth feature based method \cite{R39,R40,R41,R42,R51,R52,R52-2,R52-3} focuses on using the depth information as a supplement to color feature. Depth measure based method \cite{R43,R44,R45,R46,R47,R48,R50,R49} aims at capturing comprehensive attributes from the depth map (e.g. shape and structure) through the designed depth measures.\par

\subsection{Depth Feature Based RGBD Saliency Detection}
To achieve RGBD saliency detection, the depth feature is directly embedded into the feature pool as the supplement of color information. In \cite{R41}, color, luminance, texture, and depth features were extracted from the RGBD images to calculate the feature contrast maps. Then, the fusion and enhancement schemes were utilized to produce the final 3D saliency map. In \cite{R51}, multi-level features were used to generate the various saliency measures at different scales, then the discriminative saliency fusion strategy was designed to fuse the multiple saliency maps and obtain the final saliency result. Moreover, a bootstrap learning based salient object segmentation method was proposed. In addition, inspired by the observation that the salient regions are distinctly different from the local or global backgrounds in the depth map, ``depth contrast'' was calculated as a common depth property. In \cite{R39}, global depth contrast and domain knowledge were calculated to measure the stereo saliency. Peng \emph{et al.} \cite{R42} calculated the depth saliency through a multi-contextual contrast model, which considers the contrast prior, global distinctiveness, and background cue of depth map. Moreover, a multi-stage RGBD saliency model combining the low-level feature contrast, mid-level region grouping, and high-level prior enhancement was proposed. \par

Recently, deep learning is also successfully applied to RGBD saliency detection \cite{R52,R52-2,R52-3}. Qu \emph{et al.} \cite{R52} designed a CNN to automatically learn the interaction between low-level cues and saliency result for RGBD saliency detection. The local contrast, global contrast, background prior, and spatial prior were combined to generate the raw saliency feature vectors, which are embedded into a CNN to produce the initial saliency map. Finally, Laplacian propagation was introduced to further refine the initial saliency map and obtain the final saliency result. In addition to the multi-modal fusion problem that previous RGBD salient object detection focus on, Han \emph{et al.} \cite{R52-2} firstly exposed the cross-modal discrepancy in the RGBD data and proposed two cross-modal transfer learning strategies to better explore modal-specific representations in the depth modality. This work is the pioneering one that involves the cross-modal transfer learning problem in RGBD salient object detection. In \cite{R52-3}, Chen \emph{et al.} innovatively modelled the cross-modal complementary part including the RGB and depth data as a residual function for RGBD saliency detection. Such a re-formulation elegantly posed the problem of exploiting cross-modal complementarity as approximating the residual, making the multi-modal fusion network to be really complementarity-aware. In this work, the high-level contexts and low-level spatial cues were well-integrated, and the saliency maps were enhanced progressively. \par

\subsection{Depth Measure Based RGBD Saliency Detection}
In order to capture the comprehensive and implicit attributes from the depth map and enhance the identification of salient object, some depth measures, such as anisotropic center-surround difference measure \cite{R44}, local background enclosure measure \cite{R46}, and depth contrast increased measure \cite{R48}, are designed in different methods. Ju \emph{et al.} \cite{R44} proposed an Anisotropic Center-Surround Difference (ACSD) measure with 3D spatial prior refinement to calculate the depth-aware saliency map. Combining the ACSD measure with color saliency map, Guo \emph{et al.} \cite{R45} proposed an iterative propagation method to optimize the initial saliency map and generate the final result. Since the backgrounds always contain the regions that are highly variable in depth map, some high contrast background regions may induce false positives. To overcome this problem, Feng \emph{et al.} \cite{R46} proposed a Local Background Enclosure (LBE) measure to directly capture salient structure from depth map, which quantifies the proportion of object boundary located in front of the background. The salient objects are always placed at different depth levels and occupy small areas according to the domain knowledge in photography. Based on this observation, Sheng \emph{et al.} \cite{R48} proposed a depth contrast increased measure to pop-out the salient object through increasing the depth contrast between the salient object and distractors. Wang \emph{et al.} \cite{R50} proposed a multistage salient object detection framework for RGBD images via Minimum Barrier Distance (MBD) transform and multilayer cellular automata based saliency fusion. The depth-induced saliency map was generated through the FastMBD method, and the depth bias and 3D spatial prior were used to fuse different saliency maps at multiple stages. \par

\begin{figure}[!t]
\centering
\includegraphics[width=1\linewidth]{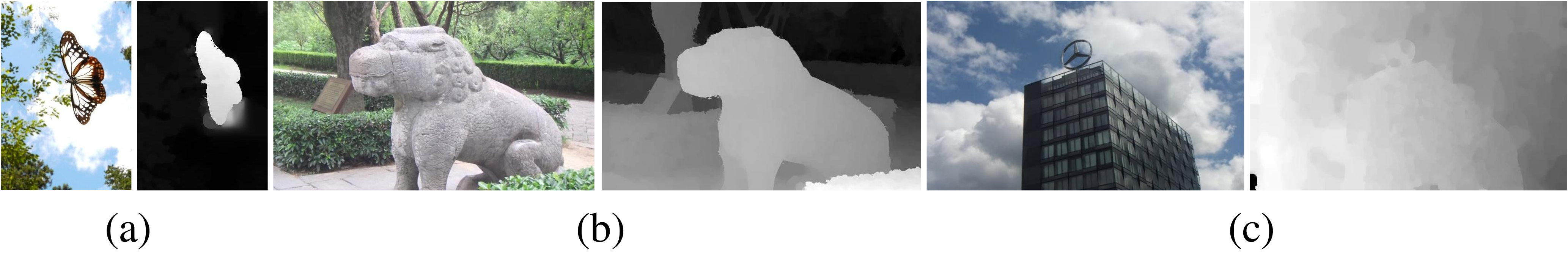}
\caption{Different quality depth maps. (a) Good depth map, $\lambda_d = 0.8014$. (b) Common depth map, $\lambda_d = 0.3890$. (c) Poor depth map, $\lambda_d = 0.0422$.}
\label{fig4}
\end{figure}

\subsection{Discussion}
Depth feature based method is an intuitive and explicit way to achieve RGBD saliency detection, which uses the depth information as an additional feature to supplement color feature, but ignores the potential attributes (e.g., shape and contour) in the depth map. By contrast, depth measure based method aims at exploiting these implicit information to refine the saliency result. However, how to effectively exploit the depth information to enhance the identification of salient object is a relatively difficult work. In addition, limited by the depth imaging techniques, sometimes the quality of the depth map is not satisfactory, as shown in Fig. \ref{fig4}(c). A good depth map benefits for the saliency detection, whereas a poor depth map may degenerate the saliency measurement.  Therefore, Cong \emph{et al.} \cite{R49} proposed a depth confidence measure to evaluate the quality of depth map, which works as a controller to constrain the introduction of depth information in the saliency model. The depth confidence measure $\lambda_d$ is defined as follows:
\begin{equation}
\lambda_d=\exp ((1-m_d)\cdot CV \cdot H)-1
\end{equation}
where $m_d$  is the mean value of depth map, $CV$ denotes the coefficient of variation, and $H$ represents the depth frequency entropy, which describes the randomness of depth distribution. With the depth confidence measure, the RGBD saliency map was generated by combining the depth-aware compactness saliency and depth-guided foreground saliency. Fig. \ref{fig4} illustrates some different quality depth maps. As visible, the depth confidence measure $\lambda_d$ effectively distinguishes different quality depth maps according to the statistical characteristics of depth map.\par

\section{Co-saliency Detection}
Co-saliency detection aims at detecting the common and salient regions from an image group containing multiple related images, which has been applied in foreground co-segmentation \cite{R54}, object co-detection \cite{R54-1}, and image matching \cite{R54-2}. In co-saliency detection, the inter-image correspondence is introduced as the common attribute constraint to discriminate the common objects from all the salient objects. To achieve co-saliency detection, some low-level or high-level features are firstly extracted to represent each image unit (e.g., superpixels), where the low-level feature describes the heuristic characteristics (e.g., color, texture, luminance), and the high-level feature captures the semantic attributes through some deep networks. Then, using these features, intra and inter saliency models are designed to explore the saliency representation from the perspectives of the individual image and inter image, respectively. For inter-image constraints capturing, different techniques are introduced, such as clustering, similarity matching, low rank analysis, and propagation. Finally, fusion and optimization schemes are utilized to generate the final co-saliency map.\par

We discuss two categories of co-saliency detection methods according to the different data, i.e., RGB co-saliency detection and RGBD co-saliency detection. Obviously, different from the RGB co-saliency detection, RGBD co-saliency detection model needs to combine the depth constraint with inter-image correspondence jointly. In addition, similar to the RGBD saliency detection, the depth cue can be used as an additional feature or a measure in RGBD co-saliency detection methods.\par

\subsection{RGB Co-saliency Detection}
As mentioned earlier, the inter-image correspondence plays an important role in co-saliency detection. In this subsection, we review some RGB co-saliency detection models based on different correspondence capturing strategies, i.e, matching based method \cite{R53,R54,R55,R57,R58,R60,R61,R132}, clustering based method \cite{R56}, rank analysis based method \cite{R59,R62}, propagation based method \cite{R63,R64}, and learning based method \cite{R65,R66,R67,R68}. A brief summary is presented in Table \ref{tab1-2}.\par

\textbf{Similarity Matching.} In most of the existing methods, the inter-image correspondence is simulated as a similarity matching process among basic units. As a pioneering work, Li and Ngan \cite{R53} proposed a co-saliency detection model for an image pair, where the inter-image correspondence is formulated as the similarity between two nodes through the normalized single-pair SimRank on a co-multilayer graph. However, this method is only applicable to image pairs. Tan \emph{et al.} \cite{R55} proposed a self-contained co-saliency detection model based on affinity matrix, which evaluates the co-saliency according to the bipartite superpixel-level graph matching across image pairs. Li \emph{et al.} \cite{R57} combined the intra and inter saliency maps to achieve co-saliency detection, where the inter-image corresponding relationship is measured by pairwise similarity ranking with pyramid features and minimum spanning tree image matching. Liu \emph{et al.} \cite{R60} proposed a hierarchical segmentation based co-saliency detection model, where the inter-image correspondence is formulated as the global similarity of each region. Li \emph{et al.} \cite{R61} proposed a saliency-guided co-saliency detection method, where the first stage recovers the co-salient parts missing in the single saliency map through the efficient manifold ranking, and the second stage captures the corresponding relationship via a ranking scheme with different queries. This model can make the existing saliency models work well in co-saliency scenarios. \par

\textbf{Clustering} is an effective way to build the inter-image correspondence, where the co-salient regions should be assigned to the same category. A cluster-based co-saliency detection algorithm without heavy learning for multiple images was proposed in \cite{R56}. Taking the cluster as the basic unit, an inter-image clustering model was designed to represent the multi-image relationship by integrating the contrast, spatial, and corresponding cues. The proposed method achieved a substantial improvement in efficiency.\par

\newcommand{\tabincell}[2]{\begin{tabular}{@{}#1@{}}#2\end{tabular}}
\begin{table}[!t]
\renewcommand\arraystretch{1.3}
\caption{Brief Introduction of RGB Co-saliency Detection}
\begin{center}
\scriptsize
\begin{tabular}{c|c|c|c}
\toprule[2.2pt]
Model & \tabincell{c}{Year} & \tabincell{c}{Inter-image\\ capturing} & Main technique \\
\hline
CSP \cite{R53} & 2011 & matching & \tabincell{c}{normalized single-pair SimRank} \\
\hline
UEM \cite{R54} & 2011 & matching & \tabincell{c}{repeatedness representation} \\
\hline
SA \cite{R55} & 2013 & matching & \tabincell{c}{superpixel-level graph matching} \\
\hline
CSM \cite{R57} & 2013 & matching &  \tabincell{c}{similarity ranking and matching} \\
\hline
RFPR \cite{R58} & 2014 & matching & inter-region dissimilarity \\
\hline
HSCS \cite{R60} & 2014 & matching & global similarity \\
\hline
SCS \cite{R61} & 2015 & matching & ranking scheme \\
\hline
HCM \cite{R132} & 2018 & matching & hierarchical consistency measure \\
\hline
CCS \cite{R56} & 2013 & clustering & \tabincell{c}{clustering with multiple cues} \\
\hline
SAW \cite{R59} & 2014 & rank analysis & rank one constraint \\
\hline
LRMF \cite{R62} & 2015 & rank analysis & \tabincell{c}{multiscale low-rank fusion} \\
\hline
CSP \cite{R63} & 2016 & propagation & two-stage propagation \\
\hline
CFR \cite{R64} & 2017 & propagation & color feature reinforcement \\
\hline
LDW \cite{R65} & 2015 & learning & deep learning, Bayesian \\
\hline
GCS \cite{R66} & 2017 & learning &  FCN framework, end-to-end \\
\hline
SPMI \cite{R67} & 2015 & learning & \tabincell{c}{self-paced multi-instance learning} \\
\hline
UML \cite{R68} & 2017 & learning & metric learning \\
\bottomrule[2.2pt]
\end{tabular}
\end{center}
\label{tab1-2}
\end{table}

\textbf{Rank Analysis.} Ideally, feature representations of co-salient objects should be similar and consistent, thus, the rank of feature matrix should appear low. Cao \emph{et al.} \cite{R59} proposed a fusion framework for co-saliency detection based on rank constraint, which is valid for multiple images and also works well on single image saliency detection. The self-adaptive weights for fusion process were determined by the low-rank energy. Moreover, this method can be used as a universal fusion framework for multiple saliency maps. Huang \emph{et al.} \cite{R62} proposed a multiscale low-rank saliency fusion method for single image saliency detection, and the Gaussian Mixture Model (GMM) was used to generate the co-saliency map via a co-saliency prior.\par

\textbf{Propagation scheme} among multiple images is presented to capture the inter-image relationship. Ge \emph{et al.} \cite{R63} proposed a co-saliency detection method based on two-stage propagation, where the inter-saliency propagation stage is utilized to discover common properties and generate the pairwise common foreground cue maps, and the intra-saliency propagation stage aims at further suppressing the backgrounds and refining the inter-saliency propagation maps. Based on the observation that co-salient objects appear similar color distributions in an abundant color feature space, Huang \emph{et al.} \cite{R64} proposed a co-saliency detection method without single saliency residuals by using color feature reinforcement. In this method, eight color features with four exponents were formed into an abundant color feature space, and co-saliency indication maps were obtained with the help of feature coding coefficients and salient foreground dictionary.\par


\textbf{Learning Model.} Recently, learning based methods for RGB co-saliency detection attract more and more attention and achieve competitive performance.\par

Deep learning has been demonstrated to be powerful in learning the high-level semantic representation, and some heuristic studies of co-saliency detection based on deep learning have been proposed. Zhang \emph{et al.} \cite{R65} proposed a co-saliency detection model from deep and wide perspectives under the Bayesian framework. From the deep perspective, some higher-level features extracted by the convolutional neural network with additional adaptive layers were used to explore better representations. From the wide perspective, some visually similar neighbors were introduced to effectively suppress the common background regions. This method is a pioneering work to achieve co-saliency detection by using deep learning, which mainly uses the convolutional network to extract better feature representations of the target. With the FCN framework, Wei \emph{et al.} \cite{R66} proposed an end-to-end group-wise deep co-saliency detection model. First, the semantic block with 13 convolutional layers was utilized to obtain the basic feature representation. Then, the group-wise feature representation and single feature representation were captured to represent the group-wise interaction information and individual image information, respectively. Finally, the collaborative learning structure with the convolution-deconvolution model was used to output the co-saliency map. The overall performance of this method is satisfactory, but the boundary of the target needs to be sharper. \par

The Multi-Instance Learning (MIL) model aims to learn a predictor for each instance through maximizing inter-class distances and minimizing intra-class distances. The Self-Paced Learning (SPL) theory is to gradually learn from the easy/faithful samples to more complex/confusable ones. Integrating the MIL regime into SPL paradigm, Zhang \emph{et al.} \cite{R67} proposed a novel framework for co-saliency detection.\par

Metric learning works on learning a distance metric to make the same-class samples closer and different-class samples as far as possible. Han \emph{et al.} \cite{R68} introduced metric learning into co-saliency detection, which jointly learns discriminative feature representation and co-salient object detector via a new objective function. This method has the capacity to handle the wide variation in image scene and achieves superior performance.\par

\subsection{RGBD Co-saliency Detection}
The superiority of depth cue has been proved in RGBD saliency detection. Combining the depth cue with inter-image correspondence, RGBD co-saliency detection can be achieved. For this task, there are two commonly used datasets, i.e., RGBD Coseg183 dataset \cite{R69} and RGBD Cosal150 dataset \cite{R71}. Limited by the data sources, only a few of methods are proposed to achieve RGBD co-saliency detection.\par

\textbf{Clustering with Depth Feature.} In \cite{R70}, Song \emph{et al.} proposed a bagging-based clustering method for RGBD co-saliency detection. The inter-image correspondence was explored via feature bagging and regional clustering. Moreover, three depth cues, including average depth value, depth range, and the Histogram of Oriented Gradient (HOG) on the depth map, were extracted to represent the depth attributes of each region. \par

\textbf{Similarity Matching with Depth Feature.} Fu \emph{et al.} \cite{R69} introduced the RGBD co-saliency map into an object-based RGBD co-segmentation model with mutex constraint, where the depth cue is utilized to enhance identification of common foreground objects and provide local features for region comparison. Introducing the depth cue as an additional feature, Cong \emph{et al.} \cite{R71} proposed a co-saliency detection method for RGBD images by using the multi-constraint feature matching and cross label propagation. The inter-image relationship was modeled at two scales, i.e., multi-constraint based superpixel-level similarity matching and hybrid feature based image-level similarity matching. Finally, cross label propagation scheme was designed to refine the intra and inter saliency maps in a cross way and generate the final co-saliency map. \par

\begin{figure}[!t]
\centering
\includegraphics[width=1\linewidth]{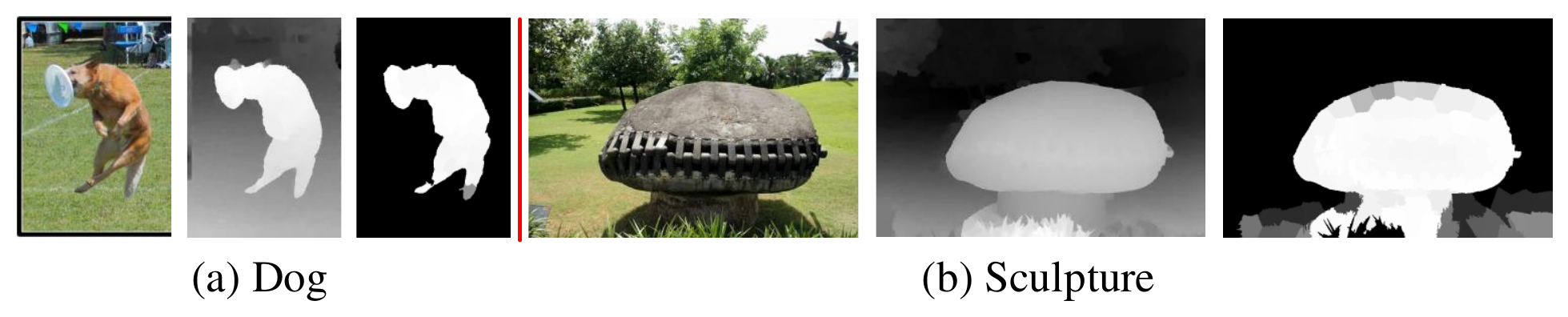}
\caption{Illustration of DSP descriptor. From the left to right in a group are the RGB image, depth map, and DSP map.}
\label{fig6}
\end{figure}

\textbf{Similarity Matching with Depth Measure.} In \cite{R72}, an iterative co-saliency detection framework for RGBD images was proposed, which integrates the addition scheme, deletion scheme, and iterative scheme. The addition scheme aimed at introducing the depth information and improving the performance of single saliency map. The deletion scheme focused on capturing the inter-image correspondence via the designed common probability function. The iterative scheme was served as an optimization process through a refinement-cycle to further improve the performance. Notably, a novel depth descriptor, named Depth Shape Prior (DSP), was designed to exploit the depth shape attribute and convert the RGB saliency map into RGBD scenarios. Fig. \ref{fig6} provides an illustration of DSP descriptor, which effectively describes the shape of salient object from the depth map. In other words, any RGB saliency map can be converted to an RGBD saliency map by using the DSP descriptor.\par

\subsection{Discussions}
Compared with image saliency detection, co-saliency detection is still an emerging topic, where the inter-image correspondence is crucial to represent the common attribute. The accurate inter-image constraint can effectively eliminate non-common saliency interference and improve the accuracy. On the contrary, the inaccurate inter-image correspondence, like noise, will degenerate the performance. The matching and propagation based methods usually capture relatively accurate inter-image relationship, but they are very time-consuming. In addition, the inter-image modeling among multiple images is a problem worth pondering in learning based method, and the stack strategy may be not a good choice for performance improvement. Of course, how to exploit the depth attribute to enhance the identification of co-salient object also needs to be further investigated.\par

\section{Video Saliency Detection}
Video sequences provide the sequential and motion information in addition to the color appearance, which benefit for the perception and identification of scene. The salient object in video is defined as the repeated, motion-related, and distinctive target. The repeated attribute constrains the salient object that should appear in most of the video frames. The motion-related characteristic is consistent with the human visual mechanism that the moving object attracts more attention than the static one. The distinctive property indicates the object should be prominent with respect to the background in each frame.\par

Most of the video saliency detection methods are dedicated to exploiting the low-level cues (e.g., color appearance, motion cue, and prior constraint) \cite{R73,R74,R75,R76,R77,R78,R79,R80,R81,R82,R83,R84,R85}. Only a few works focus on learning the high-level features and extracting the salient object in video through a learning network \cite{R86,R87,R89,R90,R91}. In the following, we will detail these two types of methods.\par

\subsection{Low-level Cue Based Video Saliency Detection}

\begin{figure}[!t]
\centering
\includegraphics[width=1\linewidth]{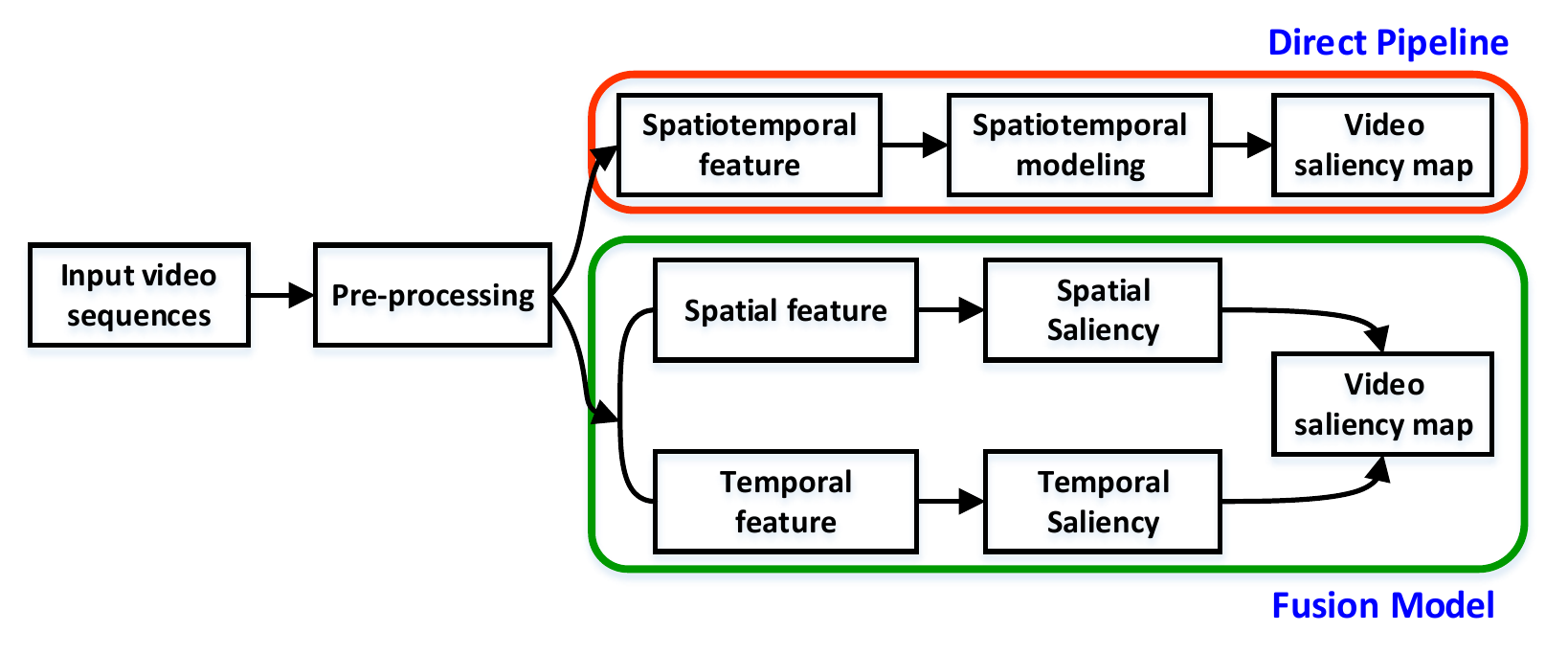}
\caption{Framework of low-level cue based video saliency detection.}
\label{fig7}
\end{figure}

According to the way of spatiotemporal extraction, low-level cue based video saliency detection method is classified into fusion model and direct-pipeline model, as shown in Fig. \ref{fig7}. For the fusion model, the spatial and temporal features are extracted to generate the spatial saliency and temporal saliency respectively, then they are combined to produce the final spatiotemporal saliency. By contrast, the direct-pipeline model directly extracts the spatiotemporal feature to generate the final spatiotemporal saliency in a straightforward and progressive way without any branches. A brief summary of the state-of-the-art methods is presented in Table \ref{tab2-2}.\par

\textbf{Fusion model} fuses the spatial saliency and temporal saliency to achieve video saliency, where the spatial cue represents the intra-frame information in each frame and the temporal cue describes the inter-frame relationship among multiple frames.\par

For spatial saliency detection, some techniques and priors in image saliency detection can be used, such as sparse reconstruction, low-rank analysis, center-surround contrast prior, and background prior. In \cite{R73}, the sparse reconstruction was utilized to discover the regions with high center-surround contrast. Fang \emph{et al.} \cite{R75} generated the static saliency map via feature contrast in compressed domain by using the luminance, color, and texture features. In \cite{R77}, the global contrast and spatial sparsity were used to measure the spatial saliency of each superpixel. Xi \emph{et al.} \cite{R82} utilized the background prior to calculate the spatial saliency. In \cite{R83}, color contrast was used to define the color saliency. \par

For temporal saliency, the motion cue is exploited to represent the moving objects in the video. In \cite{R73}, the target patch was reconstructed by overlapping patches in neighboring frames. In \cite{R75}, the motion vectors extracted from the video bitstream were used to calculate the feature differences between DCT blocks. In \cite{R77}, the superpixel-level temporal saliency was evaluated by motion distinctiveness of motion histograms. In \cite{R82}, the SIFT flow and bidirectional consistency propagation were used to define the temporal background prior. In \cite{R83}, the motion gradient guided contrast computation was used to define the temporal saliency.\par

\begin{table}[!t]
\renewcommand\arraystretch{1.3}
\caption{Brief Summary of low-level cue based Video Saliency Detection Methods}
\begin{center}
\begin{tabular}{c|c|c|c}
\toprule[2.2pt]
Model & Year & Low-level cue & Type \\
\hline
SSR \cite{R73} & 2012 & \tabincell{c}{sparse reconstruction,\\ motion trajectory} & \multirow{10}{*}{\tabincell{c}{fusion\\ model}} \\
\cline{1-3}
VSCD \cite{R75} & 2014 & \tabincell{c}{DCT coefficients, luminance,\\ color, texture, motion features} & ~ \\
\cline{1-3}
SCUW \cite{R76} & 2014 &  \tabincell{c}{luminance, color, texture, optical\\ flow, spatiotemporal uncertainty} & ~ \\
\cline{1-3}
SS \cite{R77} & 2014 & \tabincell{c}{global contrast, spatial\\ sparsity, motion histogram} & ~ \\
\cline{1-3}
STBP \cite{R82} & 2017 & \tabincell{c}{spatiotemporal background\\ prior, SIFT flow} & ~ \\
\cline{1-3}
SFLR \cite{R83} & 2017 & \tabincell{c}{spatiotemporal gradient contrast} & ~ \\
\hline
\hline
LRSD \cite{R74} & 2012 & \tabincell{c}{ low-rank, sparse decomposition} & \multirow{12}{*}{\tabincell{c}{direct-\\pipeline\\ model}} \\
\cline{1-3}
CVS \cite{R78} & 2015 & \tabincell{c}{gradient flow field,\\ local and global contrasts} & ~ \\
\cline{1-3}
RWRV \cite{R79} & 2015 & \tabincell{c}{random walk with restart,\\ motion distinctiveness, temporal\\
consistency, abrupt change} & ~ \\
\cline{1-3}
SG \cite{R81} & 2018 & \tabincell{c}{spatial edge, motion\\ boundary, geodesic distance} & ~ \\
\cline{1-3}
SGSP \cite{R84} & 2017 & \tabincell{c}{global motion histogram,\\ shortest path on graph} & ~ \\
\cline{1-3}
VSOP \cite{R85} & 2017 & \tabincell{c}{object proposals, background\\ and contrast priors, optical flow} & ~ \\
\bottomrule[2.2pt]
\end{tabular}
\end{center}
\label{tab2-2}
\end{table}

In most of the fusion based models, fusion strategy is not a key issue. In \cite{R75}, a fusion scheme considering the saliency characteristic was designed. In \cite{R77}, an adaptive fusion method at the pixel level was utilized to generate the pixel-level spatiotemporal saliency map. In \cite{R82}, the spatial and temporal saliency maps were fused via a simple addition strategy. In \cite{R83}, the modeling-based saliency adjustment and low-level saliency fusion were conducted to produce the fusion result. Furthermore, the low-rank coherency guided spatial-temporal saliency diffusion and saliency boosting strategies were adopted to improve the temporal smoothness and saliency accuracy.\par

\textbf{Direct-pipeline model} directly extracts the spatiotemporal feature to generate the final spatiotemporal saliency in a straightforward and progressive way.\par

In \cite{R74}, the stacked temporal slices along X-T and Y-T planes were used to represent the spatiotemporal feature, and the motion saliency was calculated by low-rank and sparse decomposition, where the low-rank component corresponds to the background, and the sparse proportion represents the moving foreground object.\par

Optical flow and its deformations are utilized to define the spatiotemporal feature. Wang \emph{et al.} \cite{R78} presented a spatiotemporal saliency model based on gradient flow field and energy optimization, which is robust to complex scenes, various motion patterns, and diverse appearances. The gradient flow field represented the salient regions by incorporating the intra-frame and inter-frame information. Liu \emph{et al.} \cite{R84} presented a progressive pipeline for video saliency detection, including the superpixel-level graph based motion saliency, temporal propagation, and spatial propagation. The motion saliency was measured by the shortest path on the superpixel-level graph with global motion histogram feature. Guo \emph{et al.} \cite{R85} introduced a salient object detection method for video from the perspective of object proposal via a more intuitive visual saliency analysis. The salient proposals were firstly determined by spatial saliency stimuli and contrast-based motion saliency cue. Then, proposal ranking and voting schemes were conducted to screen out non-salient regions and estimate the initial saliency. Finally, temporal consistency and appearance diversity were considered to refine the initial saliency map. It is worth learning that object proposal provides a more comprehensive and high-level representation to detect the salient object. \par

In addition, motion knowledge is used to capture the spatiotemporal feature. In \cite{R79}, the random walk with restart was exploited to detect the salient object in video, where the temporal saliency calculated by motion distinctiveness, temporal consistency, and abrupt change is employed as the restarting distribution of random walker. In \cite{R80,R81}, the spatial edge and motion boundary were incorporated as the spatiotemporal edge probability cue to estimate the initial object on the intra-frame graph, and the spatiotemporal saliency was calculated by the geodesic distance on the inter-frame graph.\par

\begin{table*}[!t]
\renewcommand\arraystretch{1.5}
\caption{A Summary of Saliency Detection Models with Comprehensive Information}
\label{tab0}
\begin{center}
\begin{tabular}{c|c|c|c}
\toprule[2.2pt]
Model & Category & Key points & Descriptions \\
\hline
\multirow{3}{*}{\tabincell{c}{RGBD\\ saliency}} & \tabincell{c}{depth\\ feature} & \tabincell{c}{use as an additional feature,\\ a heuristic and intuitive strategy} & \tabincell{c}{ directly embed into the feature pool as the supplement of color information\\ or calculate the ``depth contrast'' as a common depth property \cite{R39,R40,R41,R42,R51,R52,R52-2,R52-3}} \\
\cline{2-4}
 & \tabincell{c}{depth\\ measure} & \tabincell{c}{design a depth measure to capture\\ the shape and structure attribute} & \tabincell{c}{design some depth measures to fully exploit the effective\\ information from depth map (e.g., shape and structure) \cite{R43,R44,R45,R46,R47,R48,R49,R50}} \\
\hline
\hline
\multirow{3}{*}{\tabincell{c}{Co-\\saliency}} & \tabincell{c}{RGB\\ co-saliency} & \tabincell{c}{inter-image constraint, low-level appear-\\ance and
 high-level semantic features} & \tabincell{c}{ inter-image correspondence can be modelled as a similarity matching,\\ clustering, rank analysis, propagation, or learning process \cite{R53,R54,R55,R56,R57,R58,R59,R60,R61,R62,R63,R64,R65,R66,R67,R68,R132}} \\
\cline{2-4}
 & \tabincell{c}{RGBD\\ co-saliency} & \tabincell{c}{a new topic, inter-image constraint,\\ depth attribute (feature or measure)} & \tabincell{c}{ depth cue as a feature or a measure to enhance identification performance,\\ combine the depth cue with inter-image correspondence \cite{R69,R70,R71,R72}}\\
\hline
\hline
\multirow{3}{*}{\tabincell{c}{Video\\ saliency}} & \tabincell{c}{low-level\\ cue based} & \tabincell{c}{ direct pipeline: spatiotemporal feature\\ fusion model: spatial, temporal saliencies} & \tabincell{c}{ explore the inter-frame constraint and motion information (e.g., optical flow),\\ fusion model is more intuitive and popular \cite{R73,R74,R75,R76,R77,R78,R79,R80,R81,R82,R83,R84,R85}}\\
\cline{2-4}
 &  \tabincell{c}{learning\\ based} & \tabincell{c}{ unsupervised: stacked autoencoder\\ supervised: symmetrical deep structure} & \tabincell{c}{ learn to the spatiotemporal features and achieve competitive performance,\\ multiple frames or optical flow is embedded to represent motion cue \cite{R86,R87,R89,R90,R91}}\\
\bottomrule[2.2pt]
\end{tabular}
\end{center}
\end{table*}

In summary, the fusion model is a more intuitive method compared with the direct-pipeline model. Moreover, the existing image saliency methods can be directly used to compute the spatial saliency, which lays the foundation for spatiotemporal saliency calculation. Therefore, most of the methods pay more attention to this type.\par

\subsection{Learning Based Video Saliency Detection}
Recently, learning based video saliency detection has achieved more competitive performance, which can be roughly divided into supervised learning method \cite{R87,R89,R90} and unsupervised learning method \cite{R91}.\par

The supervised learning method aims at learning the spatiotemporal features for video saliency detection by means of a large number of labelled video sequences. Le \emph{et al.} \cite{R87} proposed a deep model to capture the SpatioTemporal deep Feature (STF), which consists of the local feature produced by a region-based CNN and the global feature computed from a block-based CNN with temporal-segments embedding. Using the STF feature, random forest and spatiotemporal conditional random field models were introduced to obtain the final saliency map. Wang \emph{et al.} \cite{R89} designed a deep saliency detection model for video, which captures the spatial and temporal saliency information simultaneously. The static network generated the static saliency map for each individual frame via the FCNs, and the dynamic network employed frame pairs and static saliency map as input to obtain the dynamic saliency result. It is worth mentioning that, a video augmentation technique was proposed to generate the labeled video training data from the existing annotated image datasets, which effectively alleviates the problem of insufficient training samples. \par

Most of the deep learning based video saliency detection methods focus on designing a separated network rather than an end-to-end network. In \cite{R90}, Le \emph{et al.} firstly proposed an \emph{\textbf{end-to-end}} 3D fully convolutional network for salient object detection in video. The Deeply Supervised 3D Recurrent Fully Convolutional Network (DSRFCN3D) contained an encoder network and a decoder network. The encoder network was used to extract the 3D deep feature from the input video block, and the decoder network aimed at computing the accurate saliency voxel. Moreover, a refinement mechanism with skip-connection and 3D Recurrent Convolution Layer (RCL3D) was designed to learn the contextual information. The loss function combined the saliency prediction loss and 3D deconvolution loss jointly, which is represented as:
\begin{equation}
L(\theta,\omega)=\zeta_{pred}(\theta,\omega_{pred})+\sum_{m=1}^{M} \zeta_{dec3D}(\theta,\omega_{dec3D}^{m})
\end{equation}
where $\theta$ is the overall network parameters, $\omega_{pred}$ denotes the weights of saliency prediction network , $\omega_{dec3D}$ represents the weights of 3D deconvolution network, $M$ is the number of 3D deconvolution layers, and $\zeta$ denotes the binary cross-entropy function.\par

Compared with supervised learning methods, only a few works focus on unsupervised learning model. As a pioneering work, Li \emph{et al.} \cite{R91} proposed an \emph{\textbf{unsupervised}} approach for video salient object detection by using the saliency-guided stacked autoencoders. First, saliency cues extracted from the spatiotemporal neighbors at three levels (i.e., pixel, superpixel, and object levels) were combined as a high-dimensional feature vector. Then, the stacked autoencoders were learned in an unsupervised manner to obtain the initial saliency map. Finally, some post-processing operations were used to further highlight the salient objects and suppress the distractors. In this method, manual intervention will be further reduced if the hand-crafted saliency cues are automatically learned from the network.\par

\subsection{Discussions}
For video saliency detection, motion cue is crucial to suppress the backgrounds and static salient objects, especially in the case of multiple objects. In general, optical flow is a common technique to represent the motion attribute. However, it is time-consuming and sometimes inaccurate, which will degenerate the efficiency and accuracy. Therefore, some deep learning based methods directly embed the continuous multiple frames into the network to learn the motion information and avoid the optical flow calculation. Of course, the video frame and optical flow can be simultaneously embedded into the network to learn the spatiotemporal feature. However, the first option may be better in terms of efficiency. In addition, the salient objects should be consistent in appearance among different frames. Therefore, some techniques, such as energy function optimization, are adopted to improve the consistency of the salient object.\par

In Table \ref{tab0}, we further summarize the characteristics of different types of saliency models, including RGBD saliency detection, co-saliency detection, and video saliency detection.\par

\section{Evaluation and Discussion}
\subsection{Evaluation Metrics}
In addition to directly comparing the saliency map with ground truth, some evaluation metrics are developed to quantitatively evaluate the performance of saliency detection methods, such as Precision-Recall (PR) curve, F-measure, Receive Operator Characteristic (ROC) curve, Area Under the Curve (AUC) score, and Mean Absolute Error (MAE). \par
\textbf{Precision-Recall (PR) curve and F-measure.} By thresholding the saliency map with a series of fixed integers from 0 to 255, the binary saliency masks are achieved. Therefore, the precision and recall scores are calculated by comparing the binary mask with the ground truth. The PR curve is drawn under different precision and recall scores, where the vertical axis denotes the precision score, and the horizontal axis corresponds to the recall score. The closer the PR curve is to the upper left, the better performance achieves. In order to comprehensively evaluate the saliency map, a weighted harmonic mean of precision and recall is defined as F-measure \cite{R92}, which is expressed as:
\begin{equation}
F_{\beta}=\frac{(1+\beta^{2})Precision\times Reall}{\beta^{2}\times Precision+ Recall}
\end{equation}
where $\beta^{2}$ is generally set to 0.3 for emphasizing the precision as suggested in \cite{R17}.\par

\begin{table*}[!t]
\renewcommand\arraystretch{1.4}
\caption{Brief Introduction of Saliency Detection for RGB Image and RGBD Images}
\begin{center}
\begin{tabular}{c|c|c|c|c|c|c|c}
\toprule[2.2pt]
Dataset & \tabincell{c}{Image\\number} & \tabincell{c}{Max\\ Resolution} & Depth attribute & Object property & \tabincell{c}{Background\\ property} & \tabincell{c}{Publish\\ year} & Best performance\\
\hline
ACSD \cite{R17} & $1000$ & $400\times400$ & $-$ & single, moderate & clean, simple & $2009$ & $F_{\beta}$: $0.94$; MAE: $0.03$ \\

ECSSD \cite{R19} & $1000$ & $400\times400$ & $-$ & single, large & clean, simple & $2012$ & $F_{\beta}$: $0.88$; MAE: $0.08$ \\

DUT-OMRON \cite{R93} & $5168$ & $400\times400$ & $-$ & single, small & complex & $2013$ & $F_{\beta}$: $0.77$; MAE: $0.07$ \\

MSRA10K \cite{R94} & $10000$ & $400\times400$ & $-$ & single, largre & clean, simple & $2014$ & $F_{\beta}$: $0.93$; MAE: $0.04$ \\

PASCAL-S \cite{R95} & $1000$ & $500\times500$ & $-$ & multiple, moderate & simple & $2014$ & $F_{\beta}$: $0.81$; MAE: $0.11$ \\

HKU-IS \cite{R96} & $850$ & $400\times400$ & $-$ & multiple, moderate & clean & $2015$ & $F_{\beta}$: $0.86$; MAE: $0.06$ \\

XPIE \cite{R97} & $4447$ & $300\times300$ & $-$ & single, moderate & complex & $2017$ & $F_{\beta}$: $0.72$; MAE: $0.12$ \\
\hline
NLPR \cite{R42} & $1000$ & $640\times640$ & Kinect capturing & single, moderate & diverse & $2014$ & $F_{\beta}$: $0.82$; AUC: $0.98$ \\

NJUD \cite{R44} & $2000$ & $600\times600$ & depth estimation & single, moderate & diverse & $2015$ & $F_{\beta}$: $0.81$; AUC: $0.98$ \\
\bottomrule[2.2pt]
\end{tabular}
\end{center}
\label{tab1}
\end{table*}

\begin{table*}[!t]
\renewcommand\arraystretch{1.4}
\caption{Brief Introduction of Co-saliency Detection Datasets}
\begin{center}
\begin{tabular}{c|c|c|c|c|c|c|c|c|c}
\toprule[2.2pt]
Dataset & \tabincell{c}{Image\\number} & \tabincell{c}{Group\\ number} &  \tabincell{c}{Group\\ size} & \tabincell{c}{Depth\\ attribute} &  Resolution & \tabincell{c}{Object\\ property} & \tabincell{c}{Background\\ property} & \tabincell{c}{Publish\\ year} & Best performance\\
\hline
MSRC \cite{R98} & $240$ & $7$ & $30$-$53$ & $-$ & $320\times210$ & complex & clean, simple & $2005$ & $F_{\beta}$: $0.84$; AUC: $0.70$ \\

iCoseg \cite{R99} & $643$ & $38$ & $4$-$42$ & $-$ & $500\times300$ & multiple & diverse & $2010$ & $F_{\beta}$: $0.85$; AUC: $0.85$ \\

Image Pair \cite{R53} & $210$ & $115$ & $2$ & $-$ & $128\times100$ & single & clustered & $2011$ & $F_{\beta}$: $0.93$; AUC: $0.97$ \\

Cosal2015 \cite{R100} & $2015$ & $50$ & $26$-$52$ & $-$ & $500\times333$ & multiple & clustered & $2016$ & $F_{\beta}$: $0.71$; AUC: $0.90$ \\

INCT2016 \cite{R101} & $291$ & $12$ & $15$-$31$ & $-$ & $500\times375$ & multiple & complex & $2016$ & $-$ \\
\hline
\tabincell{c}{RGBD\\ Coseg183 \cite{R69}} & $183$ & $16$ & $12$-$36$ & \tabincell{c}{Kinect\\ capturing} & $640\times480$ & multiple & \tabincell{c}{clustered,\\ complex} & $2015$ & $F_{\beta}$: $0.71$; MAE: $0.06$ \\

\tabincell{c}{RGBD\\ Cosal150 \cite{R71}} & $150$ & $21$ & $2$-$20$ & \tabincell{c}{depth\\ estimation} & $600\times600$ & single & diverse & $2018$ & $F_{\beta}$: $0.84$; MAE: $0.14$ \\
\bottomrule[2.2pt]
\end{tabular}
\end{center}
\label{tab2}
\end{table*}

\textbf{Receive Operator Characteristic (ROC) curve and AUC score.} The ROC curve describes the relationship between the false positive rate (FPR) and true positive rate (TPR), which is represented as:
\begin{equation}
TPR = \frac{|S_F\bigcap G_F|}{|G_F|}, FPR = \frac{|S_F\bigcap G_B|}{|G_B|}
\end{equation}
where $S_F$, $G_F$, and $S_B$ denote the set of detected foreground pixels in the binary saliency mask, the set of foreground pixels in the ground truth, and the set of background pixels in the ground truth, respectively. The closer the ROC curve is to the upper right, the better performance achieves. AUC score is the area under the ROC curve, and the larger, the better.\par
\textbf{Mean Absolute Error (MAE) score.} MAE score directly evaluates the difference between the continuous saliency map $S$ and ground truth $G$ directly:
\begin{equation}
MAE=\frac{1}{w\times h} \sum_{x=1}^{w} \sum_{y=1}^{h} |S(x,y)-G(x,y)|
\end{equation}
where $w$ and $h$ represent the width and height of the image, respectively. The smaller the MAE score is, the more similar to the ground truth, and the better performance achieves.\par
\subsection{Datasets}

\begin{table*}[!t]
\renewcommand\arraystretch{1.4}
\caption{Brief Introduction of Video Saliency Detection Datasets}
\begin{center}
\begin{tabular}{c|c|c|c|c|c|c|c|c}
\toprule[2.2pt]
Dataset & \tabincell{c}{Frame\\number} & \tabincell{c}{Video\\ number} &  \tabincell{c}{Video\\ size} &  Resolution & \tabincell{c}{Object\\ property} & \tabincell{c}{Background\\ property} & \tabincell{c}{Publish\\ year} & Best performance\\
\hline
SegTrackV1 \cite{R102} & $244$ & $6$ & $21$-$71$ & $414\times352$ & single & diverse & $2010$ & $F_{\beta}$: $0.88$; MAE: $0.10$ \\

SegTrackV2 \cite{R103} & $1065$ & $14$ & $21$-$279$ & $640\times360$ & single & diverse & $2013$ & $F_{\beta}$: $0.92$; MAE: $0.02$ \\

ViSal \cite{R78} & $963$ & $17$ & $30$-$100$ & $512\times228$ & single & diverse & $2015$ & $F_{\beta}$: $0.85$; MAE: $0.03$ \\

MCL \cite{R79} & $3689$ & $9$ & $131$-$789$ & $480\times270$ & single, small & complex & $2015$ & $-$ \\

DAVIS \cite{R104} & $3455$ & $50$ & $25$-$104$  & $1920\times1080$ & multiple & complex & $2016$ & $F_{\beta}$: $0.82$; MAE: $0.03$ \\

UVSD \cite{R84} & $6524$ & $18$ & $71$-$307$ &  $352\times288$ & single, small & clustered, complex & $2017$ & $F_{\beta}$: $0.51$; MAE: $0.10$ \\

VOS \cite{R91} & $116103$ & $200$ & $\sim500$ &  $800\times800$ & single & complex & $2018$ & $F_{\beta}$: $0.78$; MAE: $0.05$ \\
\bottomrule[2.2pt]
\end{tabular}
\end{center}
\label{tab3}
\end{table*}

In this section, we introduce the datasets for (RGBD) image saliency detection, co-saliency detection, and video saliency detection, respectively.\par

For image saliency detection, a number of datasets have been constructed over the past decade, including some large datasets with pixel-level annotations, such as DUT-OMRON \cite{R93}, MSRA10K \cite{R94}, HKU-IS \cite{R96}, and XPIE \cite{R97}, as listed in Table \ref{tab1}. Benefiting from the growth of data volume, deep learning based RGB saliency detection methods have achieved superior performance. \par 

In contrast, the datasets with pixel-wise ground truth annotations for RGBD saliency detection are relatively inadequate, which only consist of NLPR dataset \cite{R42} and NJUD dataset \cite{R44}, as listed in the last two rows of Table \ref{tab1}. The NLPR dataset includes 1000 RGBD images with the resolution of $640\times640$, where the depth maps are captured by Microsoft Kinect. The NJUD dataset is released on 2015, which includes 2000 RGBD images with the resolution of $600\times600$. The depth map in the NJUD dataset is estimated by the stereo images.\par

\begin{figure*}[!t]
\centering
\includegraphics[width=0.9\linewidth]{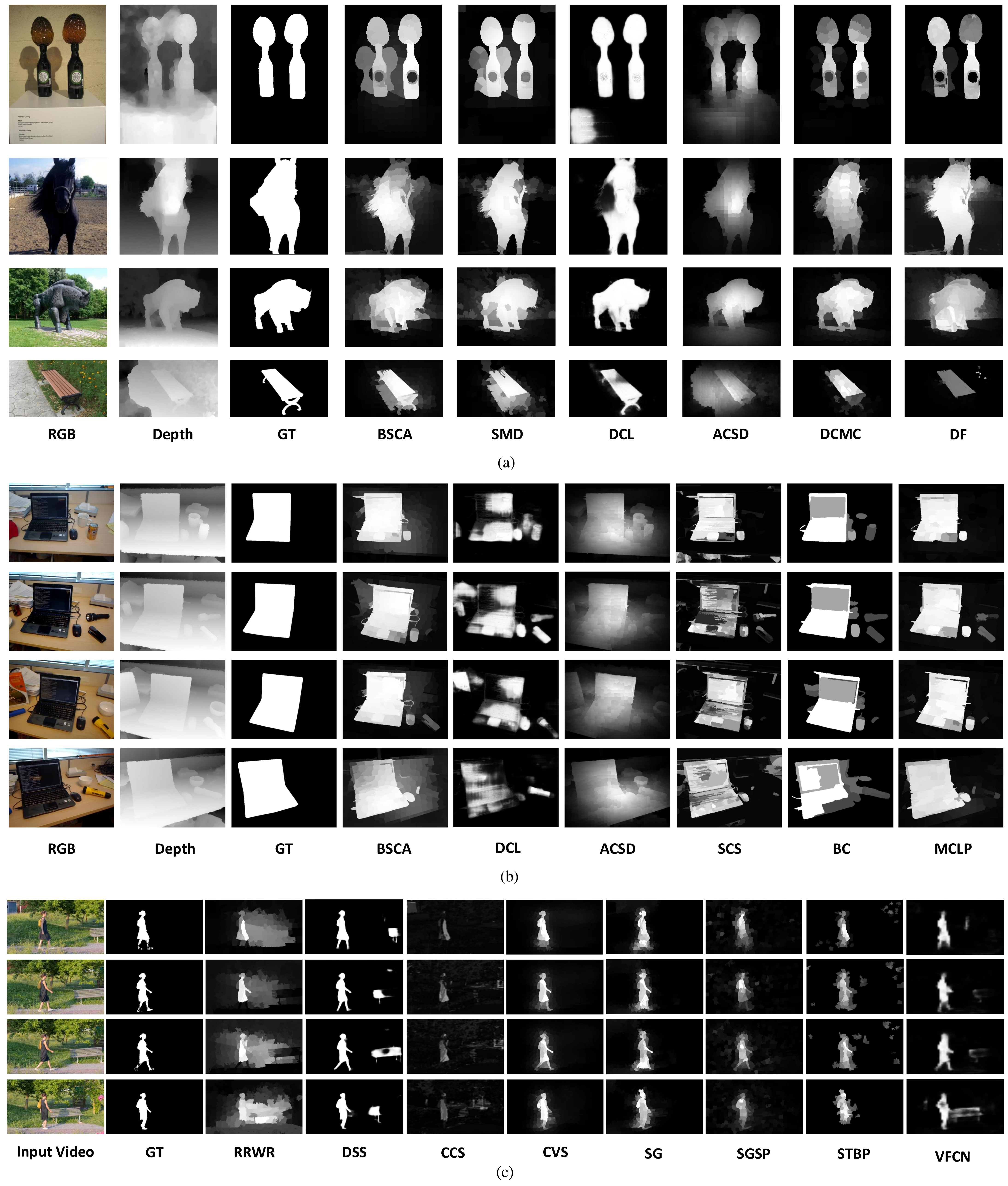}
\caption{Visual examples of different methods on different datasets. (a) NJUD dataset. (b) RGBD Coseg183 dataset. (c) DAVIS dataset.}
\label{fig9}
\end{figure*}

\begin{figure*}[!t]
\centering
\includegraphics[width=1\linewidth]{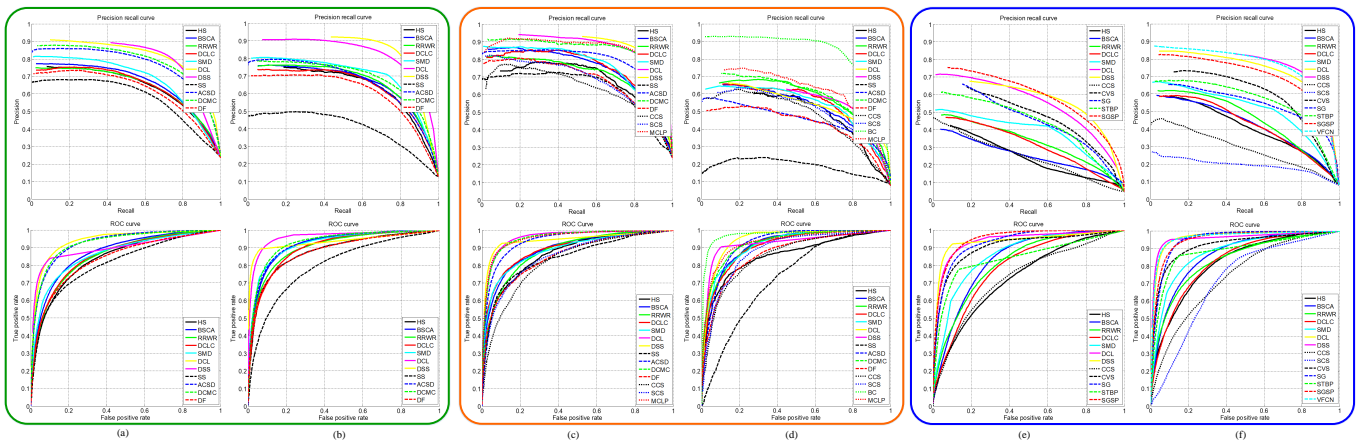}
\caption{PR and ROC curves of different methods on different datasets. (a) NJUD dataset. (b) NLPR dataset. (c) RGBD Cosal150 dataset. (d) RGBD Coseg183 dataset. (e) UVSD dataset. (f) DAVIS dataset.}
\label{fig10}
\end{figure*}

For co-saliency detection, five RGB datasets and two RGBD datasets are commonly used for evaluation, as listed in Table \ref{tab2}. MSRC \cite{R98} is a challenging dataset with complex background, which contains 7 image groups of totally 240 images with manually pixel-wise ground truth. The iCoseg \cite{R99} dataset consists of 38 image groups of totally 643 images, and the manually labeled pixel-wise ground-truth masks is also provided. Image Pair \cite{R53} dataset only contains image pairs, whereas other datasets usually include more than two images in each group. A larger co-saliency detection dataset named RGBD Cosal2015 is constructed in \cite{R100}, which consists of 2015 RGB images distributed in 50 image groups with pixel-wise ground truth. INCT2016 \cite{R101} is a more challenging dataset with larger appearance variation, indefinite number of targets, and complicated backgrounds, which contains 291 images distributed in 12 categories with pixel-level ground truth. There are two commonly used datasets with pixel-level hand annotations for RGBD co-saliency detection. One is the RGBD Coseg183 dataset \cite{R69}, which contains 183 RGBD images in total that distributed in 16 image groups. The other one is the RGBD Cosal150 dataset \cite{R71}, which collects 21 image groups containing a total of 150 RGBD images.\par

For video saliency detection, many datasets have been released, such as ViSal \cite{R78}, MCL \cite{R79}, UVSD \cite{R84}, VOS \cite{R91}, SegTrackV1 \cite{R102}, SegTrackV2 \cite{R103}, and DAVIS \cite{R104}, as listed in Table \ref{tab3}. The DAVIS dataset is a commonly used and challenging dataset, which contains 50 video sequences with the fully-annotated pixel-level ground truth for each frame. The UVSD dataset is a specially designed and newly established dataset for video saliency detection, which consists of 18 unconstrained videos with complicated motion patterns and cluttered scenes, and the pixel-wise ground truth for each frame is available. A very large video saliency detection dataset named VOS is constructed, which consists of 116103 frames in total that distributed in 200 video sequences. In this dataset, 7467 frames are annotated into binary ground truth, which is suitable for training and learning a deep model to extract the salient object in video.\par

\subsection{Comparison and Analysis}

We report some visual examples and quantitative comparisons in this section, and the related results on different datasets are shown in Figs. \ref{fig9}-\ref{fig10} and Tables \ref{tab4}-\ref{tab6}. All the results are directly provided by the authors or implemented by the source codes.\par

\begin{table}[!t]
\renewcommand\arraystretch{1.6}
\caption{Quantitative Comparisons of Different Methods on NJUD and NLPR Datasets, where ``*'' Denotes the Deep Learning Based Methods}
\label{tab4}
\begin{center}
\scriptsize
\setlength{\tabcolsep}{1.5mm}{
\begin{tabular}{c|c|c|c|c|c|c}
\toprule[2.2pt]
  & \multicolumn{3}{c|}{NJUD Dataset} & \multicolumn{3}{c}{NLPR Dataset} \\[0.5ex]
\cline{2-7}
 & $F_{\beta}$ & AUC & MAE & $F_{\beta}$ & AUC & MAE \\
\hline

HS \cite{R19} & $0.6494$ & $0.8390$ & $0.2516$ & $0.6659$ & $0.8785$ & $0.1918$ \\

BSCA \cite{R23} & $0.6672$ & $0.8709$ & $0.2148$ & $0.6702$ & $0.9207$ & $0.1768$ \\

RRWR \cite{R24} & $0.6520$ & $0.8510$ & $0.2161$ & $0.6804$ & $0.9038$ & $0.1575$ \\

DCLC \cite{R26} & $0.6527$ & $0.8526$ & $0.2007$ & $0.6662$ & $0.8992$ & $0.1381$ \\

SMD \cite{R27} & $0.6900$ & $0.8635$ & $0.1950$ & $0.7138$ & $0.9229$ & $0.1300$ \\

DCL* \cite{R33} & $0.7863$ & $0.9393$ & $0.1236$ & $0.7995$ & $0.9617$ & $0.0727$ \\

DSS* \cite{R36} & $0.7971$ & $0.8940$ & $0.1147$ & $0.8384$ & $0.9360$ & $0.0583$ \\

\hline

SS \cite{R39} & $0.6128$ & $0.8103$ & $0.2227$ & $0.4712$ & $0.8007$ & $0.1752$  \\

ACSD \cite{R44} & $0.7459$ & $0.9259$ & $0.1939$ & $0.6695$ & $0.9229$ & $0.1635$ \\

DCMC \cite{R49} & $0.7591$ & $0.9258$ & $0.1716$ & $0.6975$ & $0.9289$ & $0.1168$ \\

DF* \cite{R52} & $0.6383$ & $0.8338$ & $0.2022$ & $0.6407$ & $0.8801$ & $0.1156$ \\

\bottomrule[2.2pt]
\end{tabular}}
\end{center}
\end{table}

\begin{table}[!t]
\renewcommand\arraystretch{1.4}
\caption{Quantitative Comparisons of Different Methods on RGBD Cosal150 and RGBD Coseg183 Datasets, where ``*'' Denotes the Deep Learning Based Methods}
\label{tab5}
\begin{center}
\scriptsize
\setlength{\tabcolsep}{1.5mm}{
\begin{tabular}{c|c|c|c|c|c|c}
\toprule[2.2pt]
  & \multicolumn{3}{c|}{RGBD Cosal150 Dataset} & \multicolumn{3}{c}{RGBD Coseg183 Dataset} \\[0.5ex]
\cline{2-7}
 & $F_{\beta}$ & AUC & MAE & $F_{\beta}$ & AUC & MAE \\
\hline

HS \cite{R19} & $0.7101$ & $0.8644$ & $0.2375$ & $0.5645$ & $0.8540$ & $0.2018$ \\

BSCA \cite{R23} & $0.7318$ & $0.8914$ & $0.1925$ & $0.5678$ & $0.9164$ & $0.1877$ \\

RRWR \cite{R24} & $0.7106$ & $0.8797$ & $0.1967$ & $0.6089$ & $0.9163$ & $0.1504$ \\

DCLC \cite{R26} & $0.7385$ & $0.8913$ & $0.1728$ & $0.5994$ & $0.9073$ & $0.1097$ \\

SMD \cite{R27} & $0.7494$ & $0.8863$ & $0.1774$ & $0.5760$ & $0.9161$ & $0.1229$ \\

DCL* \cite{R33} & $0.8345$ & $0.9580$ & $0.1056$ & $0.5531$ & $0.9448$ & $0.0967$ \\

DSS* \cite{R36} & $0.8540$ & $0.9404$ & $0.0869$ & $0.5972$ & $0.9200$ & $0.0783$ \\

\hline

SS \cite{R39} & $0.6744$ & $0.8453$ & $0.2052$ & $0.2567$ & $0.7295$ & $0.1716$  \\

ACSD \cite{R44} & $0.7788$ & $0.9410$ & $0.1806$ & $0.4787$ & $0.9226$ & $0.1940$ \\

DCMC \cite{R49} & $0.8348$ & $0.9551$ & $0.1498$ & $0.6169$ & $0.9253$ & $0.1009$ \\

DF* \cite{R52} & $0.6844$ & $0.8510$ & $0.1945$ & $0.4840$ & $0.8654$ & $0.1077$ \\

\hline

CCS \cite{R56} & $0.6311$ & $0.8242$ & $0.2138$ & $0.5383$ & $0.8563$ & $0.1210$ \\

SCS \cite{R61} & $0.6724$ & $0.8515$ & $0.1966$ & $0.5553$ & $0.8797$ & $0.1616$ \\

\hline

BC \cite{R70} & $-$ & $-$ & $-$ & $0.8262$ & $0.9746$ & $0.0541$  \\
MCLP \cite{R71} & $0.8403$ & $0.9550$ & $0.1370$ & $0.6365$ & $0.9294$ & $0.0979$  \\

\bottomrule[2.2pt]
\end{tabular}}
\end{center}
\end{table}

\textbf{Image Saliency Detection vs RGBD Saliency Detection.} We evaluate the RGB saliency detection methods (HS \cite{R19}, BSCA \cite{R23}, RRWR \cite{R24}, DCLC \cite{R26}, SMD \cite{R27}, DCL \cite{R33}, and DSS \cite{R36}) and RGBD saliency detection methods (SS \cite{R39}, ACSD \cite{R44}, DCMC \cite{R49}, and DF \cite{R52}) on the NJUD and NLPR datasets, where the DCL, DSS, and DF are the deep learning based methods.\par

Fig. \ref{fig9}(a) presents some visual examples on the NJUD dataset, where the first three columns correspond to the input RGBD images and ground truth. The foreground object is effectively popped out, and the background is suppressed in the first three depth maps. In other words, the depth map can provide useful information to enhance the identification of the salient object. For the unsupervised RGB saliency detection methods (BSCA \cite{R23} and SMD \cite{R27}), some backgrounds are wrongly detected, such as the shadow in the first image and the trees in the third image. Introducing the depth cue, the consistency of salient object and the false positive in background regions are obviously improved. For example, the trees in the second and third images are effectively suppressed by the DCMC method \cite{R49}. For the supervised learning method, the DCL method \cite{R33} shows the competitive performance benefitting from the deep learning technique with a large number of labelled training data. However, limited by the annotated RGBD saliency data, the DF method \cite{R52} cannot completely suppress the backgrounds (e.g., the trees in the third image) and lose some foreground details (e.g., the chair legs in the fourth image).\par

The PR and ROC curves are shown in Fig. \ref{fig10}(a)-(b). As can be seen, on both two datasets, the deep learning based RGB saliency detection methods (DSS \cite{R36} and DCL \cite{R33}) achieve the top two performances on the PR and ROC curves, and the unsupervised RGBD saliency model (DCMC \cite{R49}) reaches the third precision on the ROC curve. Table \ref{tab4} reports the quantitative measures of different saliency methods on these two datasets, including the F-measure, AUC score, and MAE score. The overall trend of quantitative comparisons is consistent with the visualization results, that is, the performances of the unsupervised RGBD saliency models (e.g., ACSD \cite{R44} and DCMC \cite{R49}) are significantly superior to the unsupervised RGB saliency detection methods, with the maximum percentage gain of 18\% on the NJUD dataset in terms of the F-measure. Benefitting from the supervised learning with a large number of labelled data, deep learning based image saliency detection methods (DCL \cite{R33} and DSS \cite{R36}) yield the decent performance, even superior to the RGBD saliency methods. For different RGBD saliency detection methods, SS method \cite{R39} only focuses on some straightforward domain knowledge from the depth map, thus the performance is unsatisfactory. ACSD method \cite{R44} designs a novel depth measure to fully capture the depth attributes, and achieves appreciable performance. However, it does not have the ability to distinguish different quality of depth map. Introducing the depth confidence measure, DCMC method \cite{R49} is more robust to the poor depth map, and achieves more stable performance. For the deep learning based RGBD saliency detection method (i.e., DF \cite{R52}), due to the lack of labelled RGBD images, the performance is not satisfactory.\par

\begin{table}[!t]
\renewcommand\arraystretch{1.4}
\caption{Quantitative Comparisons of Different Methods on UVSD and DAVIS Datasets, where ``*'' Denotes the Deep Learning Based Methods}
\label{tab6}
\begin{center}
\scriptsize
\setlength{\tabcolsep}{1.5mm}{
\begin{tabular}{c|c|c|c|c|c|c}
\toprule[2.2pt]
  & \multicolumn{3}{c|}{UVSD Dataset} & \multicolumn{3}{c}{DAVIS Dataset} \\[0.5ex]
\cline{2-7}
 & $F_{\beta}$ & AUC & MAE & $F_{\beta}$ & AUC & MAE \\

\hline

HS \cite{R19} & $0.3258$ & $0.7288$ & $0.2727$ & $0.4552$ & $0.8170$ & $0.2495$ \\

BSCA \cite{R23} & $0.3059$ & $0.8320$ & $0.2227$ & $0.4731$ & $0.8602$ & $0.1945$ \\

RRWR \cite{R24} & $0.3931$ & $0.8152$ & $0.1842$ & $0.5138$ & $0.8328$ & $0.1678$ \\

DCLC \cite{R26} & $0.3878$ & $0.7899$ & $0.1249$ & $0.4812$ & $0.8250$ & $0.1337$ \\

SMD \cite{R27} & $0.4521$ & $0.8599$ & $0.1347$ & $0.5434$ & $0.8774$ & $0.1506$ \\

DCL* \cite{R33} & $0.5759$ & $0.9376$ & $0.0593$ & $0.7200$ & $0.9647$ & $0.0630$ \\

DSS* \cite{R36} & $0.5967$ & $0.9362$ & $0.0480$ & $0.7564$ & $0.9576$ & $0.0500$ \\

\hline

CCS \cite{R56} & $0.3124$ & $0.7303$ & $0.1107$ & $0.3485$ & $0.7418$ & $0.1506$ \\

SCS \cite{R61} & $-$ & $-$ & $-$ & $0.2305$ & $0.7166$ & $0.2569$ \\

\hline

CVS \cite{R78} & $0.5122$ & $0.9052$ & $0.1031$ & $0.6251$ & $0.9162$ & $0.0995$ \\

SG \cite{R80} & $0.4851$ & $0.9310$ & $0.1050$ & $0.5600$ & $0.9485$ & $0.1027$ \\

STBP \cite{R82} & $0.4914$ & $0.8443$ & $0.0840$ & $0.5859$ & $0.8842$ & $0.1016$ \\

SGSP \cite{R84} & $0.6016$ & $0.9505$ & $0.1585$ & $0.6944$ & $0.9504$ & $0.1375$ \\

VFCN* \cite{R89} & $-$ & $-$ & $-$ & $0.7488$ & $0.9637$ & $0.0588$ \\

\bottomrule[2.2pt]
\end{tabular}}
\end{center}
\end{table}

\textbf{Image Saliency Detection vs Co-saliency Detection.} We evaluate four types of saliency detection methods on the RGBD Cosal150 and RGBD Coseg183 datasets, including image saliency detection methods (HS \cite{R19}, BSCA \cite{R23}, RRWR \cite{R24}, DCLC \cite{R26}, SMD \cite{R27}, DCL \cite{R33}, and DSS \cite{R36}), RGBD saliency detection methods (SS \cite{R39}, ACSD \cite{R44}, DCMC \cite{R49}, and DF \cite{R52}), co-saliency detection methods (CCS \cite{R56} and SCS \cite{R61}), and RGBD co-saliency detection methods (BC \cite{R70} and MCLP \cite{R71}). \par

In Fig. \ref{fig9}(b), we present an image group with one common salient object (i.e., black computer) and cluttered backgrounds. From the figure, we can see that the image saliency detection methods (i.e., BSCA \cite{R23} and DCL \cite{R33}) cannot achieve better visual result with consistently highlighted salient objects and effectively suppressed background regions. For example, the desk with high luminance is wrongly detected by the unsupervised BSCA method \cite{R23}, and the salient objects are not effectively highlighted by the deep learning based DCL method \cite{R33} due to the complex and cluttered backgrounds. By contrast, considering the inter-image corresponding relationship, some backgrounds (e.g., the desk) are effectively suppressed by the RGB co-saliency detection method (e.g., SCS \cite{R61}). Moreover, the performance of RGBD co-saliency detection method with the depth constraint is superior to the RGB co-saliency detection method. For example, the MCLP method \cite{R71} achieves the best visual performance compared with other methods in Fig. \ref{fig9}(b). The computer in each image is highlighted more consistent and homogeneous, while the backgrounds (e.g., the desk) and non-common objects (e.g., the yellow flashlight, red hat, and orange can) are effectively eliminated. However, the flashlights with the same color as the computer are mistakenly reserved. The main reason is that, the low-level feature based method primarily focuses on capturing the color appearances from the image, while ignoring the high-level semantic attributes. The consistent conclusion can be drawn from the quantitative results in Fig. \ref{fig10}(c)-(d) and Table \ref{tab5}. On the RGBD Coseg183 dataset, compared with the deep learning based DCL method \cite{R33}, RGB co-saliency detection SCS method \cite{R61} achieves better performance in terms of F-measure. Moreover, benefitting from the depth cue and inter-image constraint, RGBD co-saliency models achieve more competitive performances. For example, the percentage gain of the BC method \cite{R70} reaches at least 33.9\% in terms of F-measure compared with others on the RGBD Coseg183 dataset, which indirectly proves the importance role of these information in co-saliency detection.\par

\begin{table}[!t]
\renewcommand\arraystretch{1.5}
\caption{Comparisons of the average running time (seconds per image) on the RGBD Cosal150 dataset}
\begin{center}
\setlength{\tabcolsep}{1.5mm}{
\begin{tabular}{c|c|c|c|c|c}
\toprule[1.2pt]
Method & DCLC \cite{R26} & SMD \cite{R27} & DF \cite{R52} & CCS \cite{R56} & MCLP \cite{R71} \\[0.5ex]
\hline
Time & $1.96$ & $7.49$ & $12.95$ & $2.65$ & $41.03$ \\
\bottomrule[1.2pt]
\end{tabular}}
\end{center}
\label{tab7}
\end{table}

For evaluating the running time, we tested the typical saliency detection methods, including single image saliency (DCLC \cite{R26} and SMD \cite{R27}), RGBD saliency (DF \cite{R52}), co-saliency (CCS \cite{R56}), and RGBD co-saliency (MCLP \cite{R71}), on a Quad Core 3.7GHz workstation with 16GB RAM. The codes are provided by the authors, which are implemented by using MATLAB 2014a. The comparisons of the average running time on the RGBD Cosal150 dataset are listed in Table \ref{tab7}. The single image saliency detection method only considers the visual information from the individual image and costs less running time. For example, DCLC method \cite{R26} only costs 1.96 seconds to process one image. As a deep learning based RGBD saliency detection method, DF method \cite{R52} takes 12.95 seconds for testing one image, which is relatively slow. Co-saliency detection algorithm needs to build the global corresponding constraint from the multiple images, thus it generally requires more computation time, especially for the matching based methods (such as MCLP \cite{R71}). Although the matching based co-saliency detection algorithm may be slower, it tends to achieve better performance. Moreover, the computations can be further accelerated on GPUs using C++.\par

\textbf{Image Saliency Detection / Co-saliency Detection vs Video Saliency Detection.} We compare the video saliency detection methods with image saliency detection and co-saliency detection models on the UVSD and DAVIS datasets, and the results are presented in Figs. \ref{fig9}-\ref{fig10} and Table \ref{tab6}. Video saliency detection is a more challenging task due to the complex motion patterns, cluttered backgrounds, and diversity spatiotemporal features. Five video saliency detection methods, including four unsupervised methods (CVS \cite{R78}, SG \cite{R80}, STBP \cite{R82}, SGSP \cite{R84}), and a deep learning based method (VFCN \cite{R89}) are used for comparison.\par

From the qualitative examples show in Fig. \ref{fig9}(c), the unsupervised image saliency detection methods cannot achieve superior performances due to the lack of temporal and motion constraints. For example, the background regions cannot be effectively suppressed by the RRWR method \cite{R24}, as shown in the third column of Fig. \ref{fig9}(c). Although the co-saliency detection model considers the inter-image relationship, it is still insufficient to fully represent the continuous inter-frame correspondence. Coupled with the lack of motion description, the salient object in video is not vigorously highlighted by the co-saliency model, such as the fifth column of Fig. \ref{fig9}(c). By contrast, video saliency detection methods achieve more satisfying performances both qualitatively and quantitatively. For example, the moving woman is highlighted homogeneously with clean background interference through the CVS method \cite{R78}.\par

From the quantitative results reported in Fig. \ref{fig10}(e)-(f) and Table \ref{tab6}, all the measurements of unsupervised video saliency models are superior to the unsupervised image saliency and co-saliency detection methods on these two datasets. For example, on the DAVIS dataset, the maximum percentage gain reaches 52.5\% in terms of F-measure. On the UVSD dataset, the F-measure can be improved from 0.3059 to 0.6016, with the percentage gain of 96.7\%. Notably, the deep learning based methods demonstrate excellent performance improvement, especially including the image saliency detection method (DCL \cite{R33} and DSS \cite{R36}). On the DAVIS dataset, the VFCN and DSS methods are comparable in performance and superior to the DCL method. Compared the visual examples of the DSS and VFCN methods in Fig. \ref{fig9}(c), the DSS method obtains more consistent salient regions, while the static object (i.e., the bench) is mistakenly reserved. By contrast, the bench can be partially suppressed by the VFCN method due to the introduction of motion cue and inter-frame continuity.\par

\textbf{Summary.} Taking the depth information as a supplementary feature of color information is a heuristic and intuitive strategy to achieve RGBD saliency detection. This type of method is easy to implement, and the performance can be improved by a reasonable saliency framework with depth cue. By contrast, the depth measure based method often yields better performance, because of it can further capture effective depth attributes from the original depth map rather than only focusing on some low-level statistical features. However, it is a tricky question, which requires researchers to gain insight into the characteristics of depth data, and to comprehensively explore the depth attributes of salient objects. In addition, there is an interesting phenomenon, the performance of deep learning based RGBD saliency detection method did not exceed the unsupervised methods due to the lack of annotated RGBD saliency training data. Thus, the data augmentation and network designing need to be further investigated for the deep learning based RGBD saliency detection method.\par

Compared with image saliency detection, co-saliency detection is a more challenging task because the multiple images need to be processed jointly. Therefore, the inter-image correspondence is crucial to determine the common attribute of salient objects and suppress the background regions. However, the inaccurate inter-image correspondence, like noise, may degenerate the detection performance, even not as good as some image saliency detection methods. Combining the inter-image constraint and depth cue, RGBD co-saliency detection is achieved, where the depth cue is utilized as an additional feature rather than a depth measure in most of the existing methods. Benefitting from the introduction of depth cue and inter-image constraint, the performance of RGBD co-saliency detection model is obviously improved, and the percentage gain reaches more than 30\% as shown in Table \ref{tab5}. At present, the research on RGBD co-saliency detection is relatively preliminary, and mainly focuses on unsupervised methods. Therefore, how to extract the depth and color features, capture the inter-image constraint relationship, and guarantee the consistency of salient regions are the research priorities of co-saliency detection in the future.\par

Taking the video sequences as some independent images, the image saliency detection models cannot obtain satisfying performance due to ignoring the inter-frame constraint and motion cue, especially the unsupervised methods. Considering the correspondence relationship between frames, the performance of co-saliency models are also disillusionary. There are two main reasons, i.e., (1) The salient objects in video are continuous in temporal axis and consistent among different frames. Thus, the inter-image correspondence in image group is not equal to the inter-frame constraint in video. (2) Motion information is essential to distinguish the salient object from the complex scene. Therefore, the temporal and motion cues should be fully utilized to highlight the salient object and suppress the backgrounds. Furthermore, the deep learning based methods have demonstrated the great superiority in performance. The continuous multiple frames or optical flow are embedded in the symmetrical network (e.g., convolution-deconvolution, encoder-decoder) to learn the spatiotemporal information and completely recover the salient regions. Although only a few of the deep learning based video saliency detection methods are available, it also points out an effort direction for future research.\par

\subsection{Applications}
\begin{figure}[!t]
\centering
\includegraphics[width=0.9\linewidth]{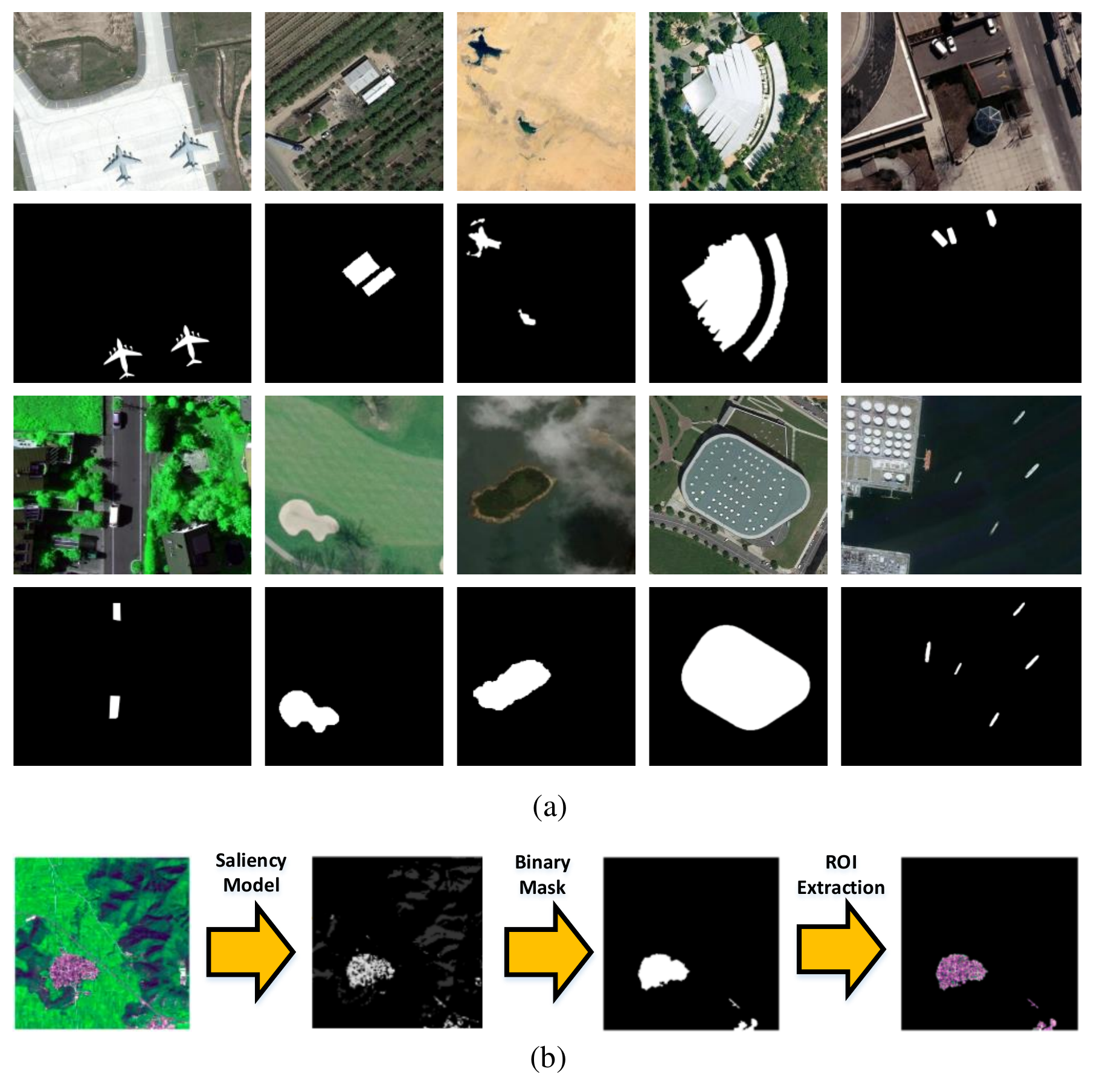}
\caption{Examples. (a) Some remote sensing images and the corresponding saliency region masks. (b) Flowchart of ROI extraction in remote sensing image.}
\label{fig11}
\end{figure}

We briefly introduce two intuitive and novel applications of saliency detection, i.e., Region-of-Interest (ROI) extraction in remote sensing image and primary object detection in video. The first task aims at extracting the ROI regions in remote sensing image, which is similar to the image saliency detection. The second one focuses on consecutively discovering the primary object in video, which is analogous to the video saliency detection.\par

\textbf{ROI Extraction in Remote Sensing Image.} With the development of imaging devices and sensors, the acquisition of high-resolution remote sensing image becomes more and more convenient and accurate. Fig. \ref{fig11}(a) shows some remote sensing images and the corresponding saliency masks. As visible, the remote sensing image is very similar to the conventional color image, except that the remote sensing image is mainly photographed from a high angle shot. Therefore, there are many small targets in remote sensing images. ROI extraction technique in remote sensing image has been applied in a variety of perception tasks, such as object detection, land cover classification, and object recognition. Generally, the saliency attribute of the object is used in most of the existing methods to constrain the ROI extraction in remote sensing image \cite{R105,R106,R107,R108,R109,R110,R111}, and the general flowchart is shown in Fig. \ref{fig11}(b). However, due to complex imaging mechanisms and different image characteristics, it is difficult to achieve satisfactory performance by directly transplanting the traditional RGB image saliency detection method to remote sensing image. There are two urgent issues that need to be addressed, i.e., (1) Extract the unique expression of salient object in remote sensing image. (2) Handle the small targets in remote sensing image. At present, this task has a very broad space for development.\par
\begin{figure}[!t]
\centering
\includegraphics[width=0.86\linewidth]{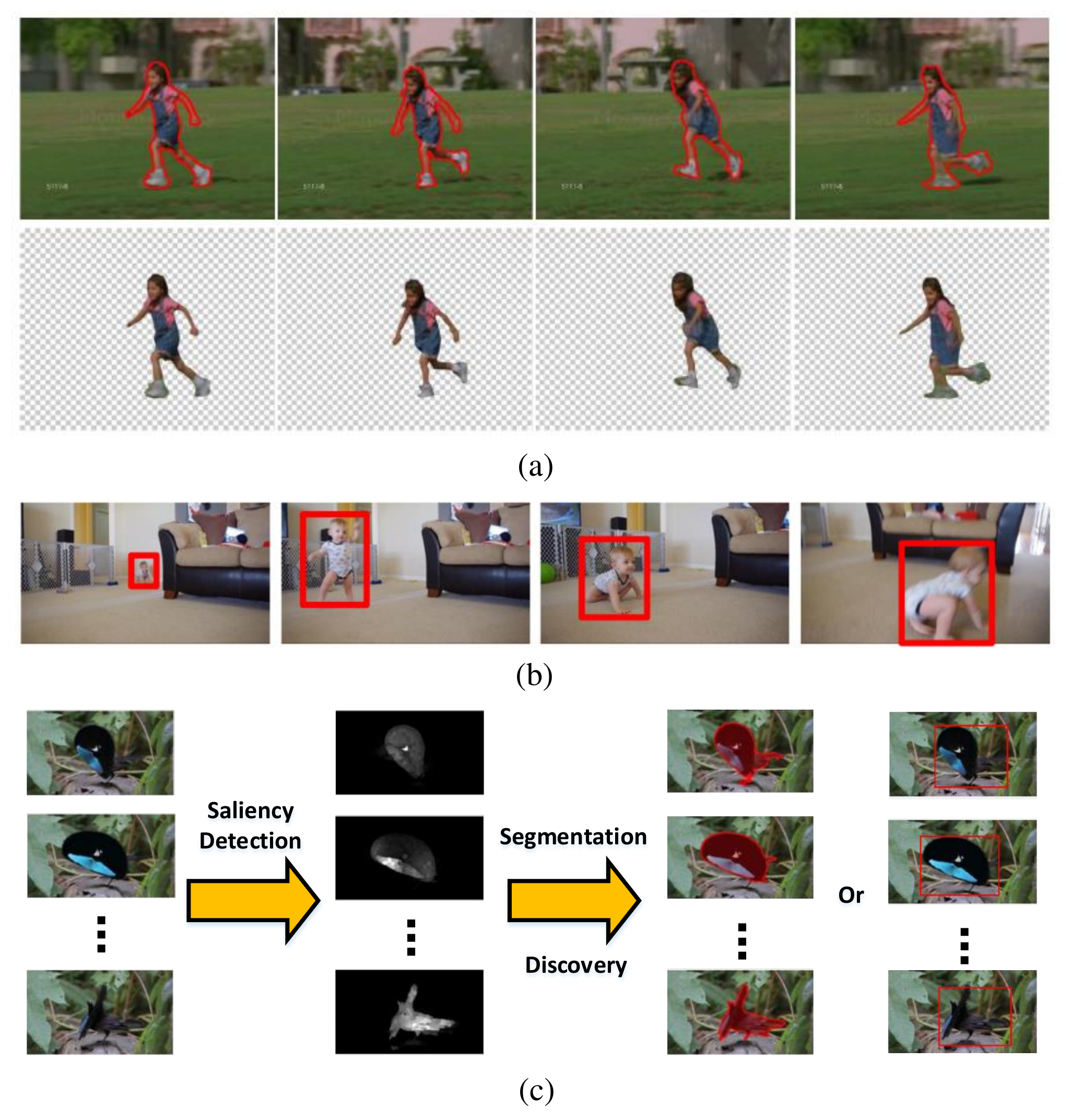}
\caption{Examples of primary object detection in video. (a) Primary object segmentation in video. (b) Primary object discovery in video. (c) Flowchart of primary object detection in video.}
\label{fig12}
\end{figure}

\textbf{Primary Object Detection in Video.} Video sequence can be divided into two components, i.e., primary objects and background regions. Primary object segmentation in video aims at obtaining the pixel-level result of the primary object, while primary object discovery locates the primary object through a bounding box, as shown in Fig. \ref{fig12}(a-b). Fig. \ref{fig12}(c) provides the flowchart of primary object detection in video, which mainly contains saliency model construction and primary object detection. In most of the existing methods, the primary object is directly defined as the salient object in video \cite{R112,R113,R114,R115,R116,R117}. In fact, there are several differences and connections between them. First, the primary object may not be the most salient one in all frames. Secondly, inter-frame correspondence works as an important temporal cue for highlighting the primary object and suppressing distractors. Thirdly, some commonly used visual priors in saliency detection may no longer be valid in video due to camera and object motion, such as background prior. Last but not least, primary object should be consistently detected from a varying scene, whereas salient object detection only considers the individual and fixed scene. Therefore, it is insufficient to only consider the image saliency attribute. The motion cue and global appearance should be introduced jointly to constrain the initial result generation. In the future, some machine learning techniques, such as deep learning and reinforcement learning, can be incorporated into the model to achieve superior performance.\par

\subsection{Challenges and Problems}

In the last decades, a plenty of saliency detection methods have been proposed to obtain the remarkable progresses and performance improvements. However, there still exist many issues that are not well resolved and needed to be further investigated in the future.\par

\textbf{For RGBD saliency detection, how to capture the accurate and effective depth representation to assist in saliency detection is a challenge.} Taking the depth information as an additional feature to supplement color feature is an intuitive and explicit way, but it ignores the potential attributes in the depth map, such as shape and contour. By contrast, depth measure based method aims at exploiting these implicit information to refine the saliency result. For example, the depth shape can be used to highlight the salient object and suppress the background, and the depth boundary can be utilized to refine the object boundary and obtain sharper saliency result. In addition, the whole object usually has high consistency in the depth map. Therefore, the depth information can be used to improve the consistency and smoothness of the acquired saliency map. Generally, depth measure based methods can achieve a better performance. However, how to effectively exploit the depth information to enhance the identification of salient object has not yet reached a consensus. On the whole, combining the explicit and implicit depth information to obtain a more comprehensive depth representation is a meaningful attempt for RGBD saliency detection.\par

\textbf{For co-saliency detection, how to explore inter-image correspondence among multiple images to constrain the common properties of salient object is a challenge.} Inter-image corresponding relationship plays an essential role in determining the common object from all the salient objects, which can be formulated as a clustering process, a matching process, a propagation process, or a learning process. However, these methods may either be noise-sensitive or time-consuming. The accuracy of corresponding relationship is directly related to the performance of the algorithm. Thus, capturing the accurate inter-image correspondence is an urgent problem to be addressed. At present, there have been some attempts to detect co-salient object using deep learning network. However, these methods often simply cascade the features produced from the single image and re-learn, rather than designing a specific inter-image network to learn the effective inter-image correspondence. \par

\textbf{For video saliency detection, how to combine more information and constraints, such as motion cue, inter-frame correspondence, and spatiotemporal consistency, is a challenge.} Motion cue plays more important role in discovering the salient object from the clustered and complex scene. The inter-frame correspondence represents the relationship among different frames, which is used to capture the common attribute of salient objects from the whole video. The spatiotemporal consistency constrains the smoothness and homogeneity of salient objects from the spatiotemporal domain. The main contributions of the existing methods are often concentrated in these three aspects. In addition, the video saliency detection algorithm based on deep learning is still immature, and only a few methods have been proposed, which is a relatively underexplored area. However, it is a challenging task to learn the comprehensive features including intra-frame, inter-frame, and motion through a deep network under the limited training samples.\par

\section{Conclusion and Future Work}
In this paper, we have conducted a survey on visual saliency detection with comprehensive information, including depth cue for RGBD saliency detection, inter-image correspondence for co-saliency detection, and temporal constraint for video saliency detection. We have reviewed the recent progress of saliency detection, analyzed the different types of saliency detection algorithms, and conducted experimental comparisons and analysis. It has been demonstrated that the performance has been improved by introducing the comprehensive information in an appropriate way. For example, the depth measure based method often yields better performance, because of it can further capture effective depth attributes from the depth map. Combining the inter-image correspondence and depth cue, RGBD co-saliency detection models achieve better performance. Through a symmetrical structure (e.g., convolution-deconvolution, encoder-decoder) with continuous multiple frames input, the deep learning based video saliency detection methods learn the high-level spatiotemporal feature and improve the efficiency.\par

In the future, some research directions and emphases of saliency detection can be focused on, i.e., (1) New attempts in learning based saliency detection methods, such as small samples training, weakly supervised learning, and cross-domain learning. Limited by the labelled training data, more work, such as designing a special network, can be explored in the future to achieve high-precision detection with small training samples. In addition, weakly supervised salient object detection method is a good choice to address the insufficient pixel-level saliency annotations. Furthermore, the cross-domain learning is another direction that needs to be addressed for learning based RGBD saliency detection method. (2) Extending the saliency detection task in different data sources, such as light filed image, RGBD video, and remote sensing image. In the light filed image, the focusness prior, multi-view information, and depth cue should be considered jointly. For the RGBD video data, the depth constraint should be introduced to assist in the spatiotemporal saliency. In the remote sensing image, due to the high angle shot photographed, some small targets and shadows are included. Thus, how to suppress the interference effectively and highlight the salient object accurately should be further investigated in the future.\par


\ifCLASSOPTIONcaptionsoff
  \newpage
\fi

{
\bibliographystyle{IEEEtran}
\bibliography{ref}
}

\begin{IEEEbiography}[{\includegraphics[width=1in,height=1.25in,clip,keepaspectratio]{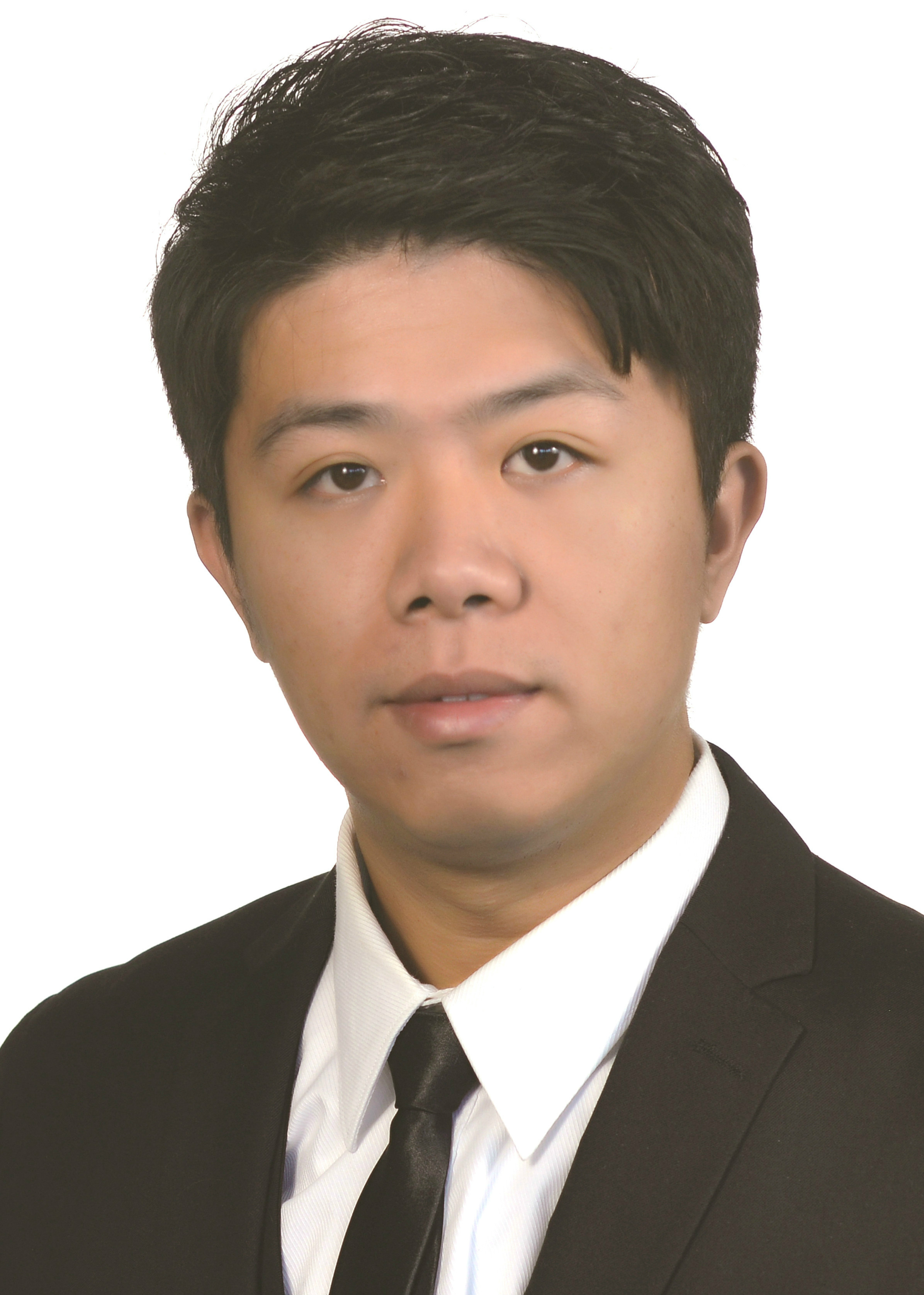}}]{Runmin Cong}
received the M.S. degree from the Civil Aviation University of China, Tianjin, China, in 2014. He is currently pursuing his Ph.D. degree in information and communication engineering with Tianjin University, Tianjin, China. \par
He was a visiting student at Nanyang Technological University (NTU), Singapore, from Dec. 2016 to Feb. 2017. Since May 2018, he has been working as a Research Associate at the Department of Computer Science, City University of Hong Kong (CityU), Hong Kong. He is a Reviewer for the IEEE TIP, TMM, and TCSVT, etc. His research interests include computer vision, image processing, saliency detection, and 3-D imaging.
\end{IEEEbiography}

\begin{IEEEbiography}[{\includegraphics[width=1in,height=1.25in,clip,keepaspectratio]{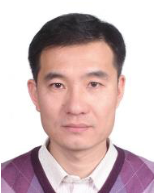}}]{Jianjun Lei}
(M'11-SM'17) received the Ph.D. degree in signal and information processing from Beijing University of Posts and Telecommunications, Beijing, China, in 2007.\par
He was a visiting researcher at the Department of Electrical Engineering, University of Washington, Seattle, WA, from August 2012 to August 2013. He is currently a Professor at Tianjin University, Tianjin, China. He is on the editorial boards of Neurocomputing and China Communications. His research interests include 3D video processing, virtual reality, and artificial intelligence.
\end{IEEEbiography}

\begin{IEEEbiography}[{\includegraphics[width=1in,height=1.25in,clip,keepaspectratio]{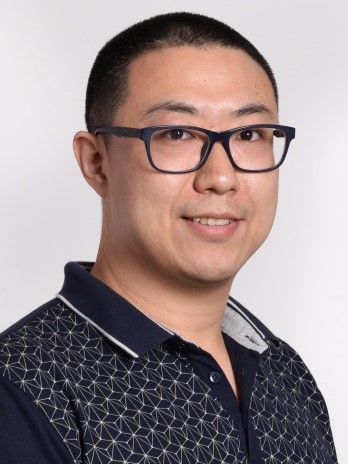}}]{Huazhu Fu}
(SM'18) received the Ph.D. degree in computer science from Tianjin University, China, in 2013. From 2013 to 2015, he worked as the research fellow at Nanyang Technological University, Singapore. And from 2015 to 2018, he worked as a Research Scientist at the Institute for Infocomm Research, Agency for Science, Technology and Research, Singapore. He is currently the Senior Scientist with the Inception Institute of Artificial Intelligence, Abu Dhabi, United Arab Emirates. His research interests include computer vision, image processing, and medical image analysis. He is the Associate Editor of IEEE Access and BMC Medical Imaging.
\end{IEEEbiography}

\vspace{-1cm}

\begin{IEEEbiography}[{\includegraphics[width=1in,height=1.25in,clip,keepaspectratio]{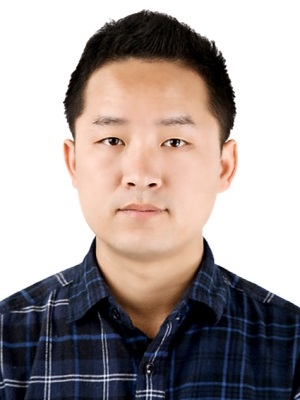}}]{Ming-Ming Cheng}
received his Ph.D. degree from Tsinghua University in 2012. Then he did 2 years research fellow, with Prof. Philip Torr in Oxford. He is now a professor at Nankai University, leading the Media Computing Lab. His research interests includes computer graphics, computer vision, and image processing. He received research awards including ACM China Rising Star Award, IBM Global SUR Award, CCF-Intel Young Faculty Researcher Program, etc.
\end{IEEEbiography}

\vspace{-1cm}

\begin{IEEEbiography}[{\includegraphics[width=1in,height=1.25in,clip,keepaspectratio]{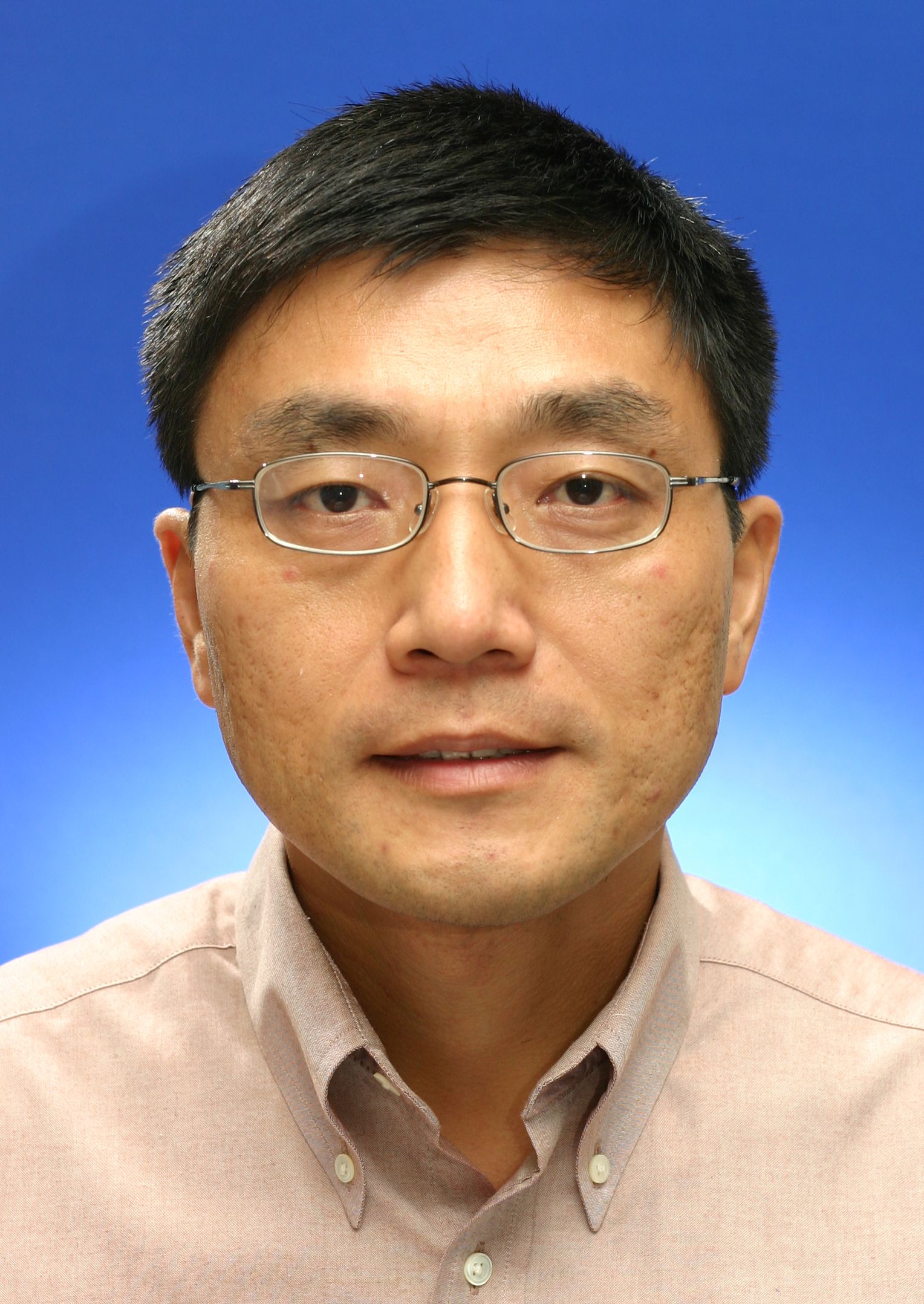}}]{Weisi Lin}
(M'92-SM'98-F'16) received his Ph.D. from King's College, London University, U.K. He is a Professor in the School of Computer Science and Engineering, Nanyang Technological University. His areas of expertise include image processing, perceptual signal modeling, video compression, and multimedia communication, in which he has published 180+ journal papers, 230+ conference papers, filed 7 patents, and authored 2 books. He has been an AE for IEEE Trans. on Image Processing, IEEE Trans. on Circuits and Systems for Video Tech., IEEE Trans. on Multimedia, and IEEE Signal Processing Letters. He has been a Technical Program Chair for IEEE ICME 2013, PCM 2012, QoMEX 2014 and IEEE VCIP 2017. He has been an invited/panelist/keynote/tutorial speaker in 20+ international conferences, as well as a Distinguished Lecturer of IEEE Circuits and Systems Society 2016-2017, and Asia-Pacific Signal and Information Processing Association (APSIPA), 2012-2013. He is a Fellow of IEEE and IET, and an Honorary Fellow of Singapore Institute of Engineering Technologists.
\end{IEEEbiography}

\vspace{-1cm}

\begin{IEEEbiography}[{\includegraphics[width=1in,height=1.25in,clip,keepaspectratio]{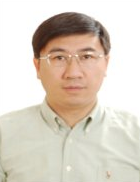}}]{Qingming Huang}
(SM'08-F'18) is a professor in the University of Chinese Academy of Sciences and an adjunct research professor in the Institute of Computing Technology, Chinese Academy of Sciences. He graduated with a Bachelor degree in Computer Science in 1988 and Ph.D. degree in Computer Engineering in 1994, both from Harbin Institute of Technology, China.\par
His research areas include multimedia video analysis, image processing, computer vision and pattern recognition He has published more than 300 academic papers in prestigious international journals including IEEE Trans. on Image Processing, IEEE Trans. on Multimedia, IEEE Trans. on Circuits and Systems for Video Tech., etc, and top-level conferences such as ACM Multimedia, ICCV, CVPR, IJCAI, VLDB, etc. He is the associate editor of IEEE Trans. on Circuits and Systems for Video Tech., and Acta Automatica Sinica, and the reviewer of various international journals including IEEE Trans. on Multimedia, IEEE Trans. on Circuits and Systems for Video Tech., IEEE Trans. on Image Processing, etc. He is a Fellow of IEEE and has served as general chair, program chair, track chair and TPC member for various conferences, including ACM Multimedia, CVPR, ICCV, ICME, PCM, PSIVT, etc.
\end{IEEEbiography}

\end{document}